\documentclass[10pt,twocolumn,letterpaper]{article}

\usepackage[pagenumbers]{cvpr} %

\usepackage{microtype}
\usepackage{multirow}
\usepackage{float}
\usepackage{colortbl}
\usepackage{tabularx}
\newcolumntype{Y}{>{\centering\arraybackslash}X}
\newcolumntype{R}{>{\raggedleft\arraybackslash}X}
\usepackage{tikz}
\usetikzlibrary{arrows.meta,positioning,fit,calc,shapes.geometric}

\definecolor{cblue}{RGB}{0,114,178}
\definecolor{cgreen}{RGB}{0,158,115}
\definecolor{cred}{RGB}{213,94,0}
\definecolor{corange}{RGB}{230,159,0}
\definecolor{cpurple}{RGB}{204,121,167}
\definecolor{cgray}{RGB}{244,246,249}
\definecolor{oursrow}{HTML}{FDF0E6}

\newcommand{\se}[1]{\ensuremath{_{\textcolor{black!50}{\pm #1}}}}

\newcommand{\method}{\textup{WARP-JEPA}}
\newcommand{\tm}{\textup{TM}}
\newcommand{\tmm}{\textup{TMM}}
\newcommand{\tmmm}{\textup{TMMM}}
\newcommand{\mm}{\textup{MM}}
\newcommand{\mmm}{\textup{MMM}}
\newcommand{\mmmm}{\textup{MMMM}}
\newcommand{\ttarch}{\textup{TT}}
\newcommand{\ttt}{\textup{TTT}}
\newcommand{\tttt}{\textup{TTTT}}

\makeatletter
\renewcommand\paragraph{\@startsection{paragraph}{4}{\z@}{1.6ex \@plus.3ex \@minus.2ex}{-1em}{\normalfont\normalsize\bfseries}}
\makeatother
\definecolor{cvprblue}{rgb}{0.21,0.49,0.74}
\usepackage[pagebackref,breaklinks,colorlinks,allcolors=cvprblue]{hyperref}

\title{WARP-JEPA: World-Model-in-the-Loop Amortized Rollout Planning\\for Frozen Hierarchical Joint-Embedding Predictive Architectures}

\author{%
Junjie Li$^{1}$ \quad Zhaojiacheng Zhou$^{1}$ \quad Kaixiang Wang$^{1}$ \quad Zihan Wang$^{1}$\\
QingSong Zhao$^{1}$ \quad Jiong Lou$^{1,*}$ \quad Jie Li$^{1,*}$\\[4pt]
$^{1}$Shanghai Jiao Tong University\\
{\small $^{*}$Corresponding authors}
}

\begin{document}
\maketitle

\begin{abstract}
Hierarchical Joint-Embedding Predictive Architectures (H-JEPA) enable long-horizon visual control but rely on costly test-time gradient descent for planning. Our analysis reveals that iterative optimization consumes up to 97.7\% of episode runtime and that low predicted rollout costs do not necessarily translate into successful control. We further identify a planning-horizon asymmetry between long-horizon top-level composition and short two-step subordinate planning. To address these challenges, we propose \textbf{\method{}}, a hierarchical amortized planning framework that replaces online gradient optimization with learned multi-candidate trajectory proposers while keeping the world models frozen. \method{} combines a lightweight world-model-in-the-loop causal Transformer at the top level with parallel residual MLPs at subordinate levels. The proposers are trained using model-based rollout costs and best-of-$K$ behavior cloning, enabling efficient candidate selection without test-time gradient optimization. Across 30 frozen H-JEPA models on four benchmarks, \method{} achieves 9.6--18.8$\times$ faster episode inference on long-horizon tasks (2.9--3.0$\times$ on Cube) and improves task success by up to 33.3 percentage points. These results demonstrate that amortized planning can substantially improve both inference efficiency and control performance without retraining the world models. Code: \url{https://github.com/hideinsewer/WARP-JEPA}.
\end{abstract}

\section{Introduction}
\label{sec:intro}

Learning world models that predict the consequences of actions is central to autonomous visual control~\cite{ha2018world,lecun2022path}. Rather than reconstructing observations pixel by pixel, Joint-Embedding Predictive Architectures (JEPAs) predict future representations in a compact latent space~\cite{assran2023self,bardes2024revisiting,lecun2022path}. Recent action-conditioned models, including DINO-WM~\cite{zhou2024dinowm} and LeWorldModel~\cite{maes2026leworldmodel}, have extended this paradigm to visual planning. Hierarchical JEPA (H-JEPA)~\cite{zhang2026hjepa} further organizes latent dynamics across multiple temporal scales, allowing high-level macro-action planning to guide lower-level control. However, H-JEPA still obtains actions by repeatedly optimizing candidate sequences through the learned world model at test time. Consequently, its control performance and inference cost depend heavily on the online planner.

\begin{figure}[t]
\centering
\includegraphics[width=\columnwidth]{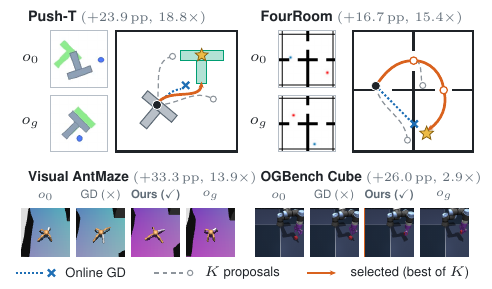}
\vspace{-4pt}
\caption{\textbf{Online gradient descent \vs \method{}} on all four benchmarks (in parentheses: success gain and speed-up of TM over Online GD, L2, \cref{tab:main}). Top: multi-modal candidate selection escapes local minima on Push-T and FourRoom. Bottom: real closed-loop rollouts of \mm{} on Visual AntMaze and OGBench Cube (full rollout strips in App.~Fig.~S1).}
\label{fig:teaser}
\vspace{-8pt}
\end{figure}

\textbf{This reliance on online optimization introduces two limitations.} First, planning is computationally expensive. Profiling 2-, 3-, and 4-level H-JEPA models across four benchmarks, we find that iterative solvers account for $96.6\%$--$97.7\%$ of episode runtime on long-horizon tasks ($79.9\%$ on the short-horizon Cube), as nested solver loops execute thousands of sequential forward-backward passes, with episodes taking up to $92.5$\,s on average on an NVIDIA H200 GPU. Second, minimizing model-predicted rollout cost does not necessarily yield reliable control. On Push-T, online gradient descent reaches a substantially lower predicted terminal cost than a behavior-regularized learned proposer, yet does not achieve higher closed-loop success and deviates further from demonstrated actions (terminal cost $16.9$ \vs $73.7$, success $42.0\%$ \vs $48.0\%$, action MSE $4.23$ \vs $1.25$; \cref{tab:mechanism}). This mismatch suggests that, beyond non-convex local minima, aggressive optimization can exploit inaccuracies in learned dynamics or drive actions away from the data-supported distribution. In hierarchical planning, inaccurate high-level subgoals can further propagate errors to subordinate levels. These observations motivate a planning mechanism that reduces online computation while favoring feasible, goal-directed actions.

\textbf{To address these limitations, we introduce \method{}} (\textbf{W}orld-model-in-the-loop \textbf{A}mortized \textbf{R}ollout \textbf{P}lanning), a hierarchical amortized planning framework for strictly frozen H-JEPA world models. Unlike online gradient descent, which directly optimizes actions against model-predicted costs, \method{} decouples online action generation from iterative gradient-based optimization through the world model. Specifically, it trains multi-candidate trajectory proposers by combining a relaxed winner-take-all world-model rollout objective with best-of-$K$ behavior cloning. This joint objective encourages goal-reaching trajectories while regularizing proposals toward demonstrated behaviors, reducing the risk of exploiting inaccuracies in the learned world model. At inference time, the learned proposers directly generate candidate trajectories, which are evaluated and selected using the frozen world model, without iterative gradient optimization for every top-level and parent-conditioned subordinate solve.

\begin{figure*}[t]
\centering
\includegraphics[width=\textwidth]{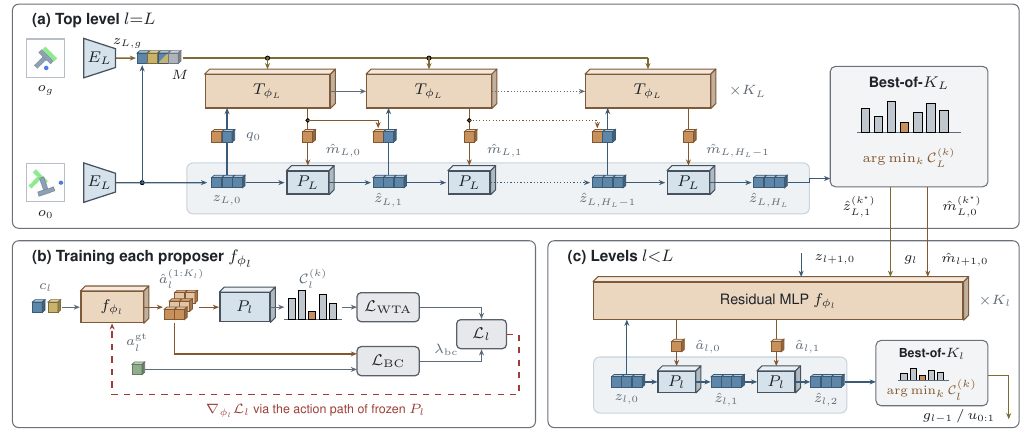}
\caption{\textbf{Architecture of \method{} (asymmetric \tm{} shown).} Blue: frozen H-JEPA modules; orange: learned proposers. \textbf{(a)} A standard 1-layer causal Transformer decoder $T_{\phi_L}$~\cite{vaswani2017attention} cross-attends to the context tokens $M$ (\cref{eq:tf_query}); each query joins the previous macro-action with the state imagined by the frozen $P_L$, which also scores the $K_L$ plans with zero extra rollouts. \textbf{(b)} Proposers are trained with relaxed WTA rollout cost plus best-of-$K$ behavior cloning (\cref{eq:loss}) through frozen $P_l$. \textbf{(c)} A residual MLP proposes $K_l$ two-step bridges in parallel, verified by one batched rollout of $P_l$.}
\label{fig:arch}
\vspace{-6pt}
\end{figure*}

\textbf{We further tailor the proposer architecture to the distinct planning requirements across hierarchy levels.} In H-JEPA, subordinate levels typically solve short, two-step bridges toward parent-specified subgoals, which can be efficiently predicted in parallel. In contrast, the top level may need to compose up to $12$ interdependent macro-actions, requiring each action to account for the intermediate states induced by previous actions. Accordingly, our default asymmetric architecture (\tm{}) employs parallel residual MLPs at subordinate levels and a lightweight one-layer causal Transformer at the top level (\cref{fig:arch}). The MLPs generate candidate trajectories in parallel, followed by a single batched world-model rollout for selection. The Transformer instead generates macro-actions autoregressively, conditioning each action on the evolving latent state predicted by the frozen world model. By integrating world-model rollouts into generation, it enables best-of-$32$ selection without additional rollouts. We also evaluate all-MLP (\mm{}) and all-Transformer (\ttarch{}) alternatives to assess the benefits of this asymmetric design.

We evaluate \method{} on 30 frozen H-JEPA models spanning two, three, and four levels across Push-T, FourRoom, Visual AntMaze, and OGBench Cube. Compared with the original online gradient-descent planner, \method{} accelerates end-to-end episode inference by $9.6$--$18.8\times$ on long-horizon tasks ($2.9$--$3.0\times$ on Cube) and improves task success by up to $33.3$ percentage points. On tasks with long top-level horizons, \tm{} further improves success over all-MLP planning by up to $8.8$ percentage points. These results demonstrate that redesigning the planner can substantially improve both efficiency and control performance without retraining the underlying world models.

Our main contributions are threefold:
\begin{itemize}
    \item \textbf{Diagnosing online planning limitations.} We quantify the dominant cost of iterative test-time optimization and identify a discrepancy between low model-predicted rollout cost and closed-loop control success, as well as a planning-horizon asymmetry between long top-level plans and two-step subordinate bridges.
    \item \textbf{Hierarchical amortized planning.} We introduce behavior-regularized, multi-candidate planning for frozen H-JEPA models, with an asymmetric architecture that combines world-model-in-the-loop top-level planning and parallel subordinate planning.
    \item \textbf{Multi-depth validation.} Across 30 frozen models and four benchmarks, we demonstrate substantial inference speedups and improved task success, without retraining the world models.
\end{itemize}

\section{Related work}
\label{sec:related}

\paragraph{Latent world models and hierarchical JEPA planning.}
Latent dynamics models support control and planning in compact learned state spaces~\cite{ha2018world,hafner2019learning,schrittwieser2020mastering}. Joint-Embedding Predictive Architectures (JEPAs)~\cite{assran2023self,bardes2024revisiting,lecun2022path,sobal2022jepaslow} predict target embeddings without a pixel-reconstruction objective. Recent action-conditioned world models plan zero-shot with sampling-based optimizers: CEM in DINO-WM~\cite{zhou2024dinowm}, V-JEPA~2-AC~\cite{assran2025vjepa2}, and LeWorldModel~\cite{maes2026leworldmodel}, and MPPI~\cite{williams2017information} in PLDM~\cite{sobal2025pldm}; gradient-based action optimization~\cite{bharadhwaj2020model} is an alternative, and random-shooting MPC~\cite{nagabandi2018neural} an early learned-model baseline. To scale to long horizons, hierarchical world models decompose control via visual subgoal generation~\cite{nair2020hierarchical}, reward-driven manager-worker policies~\cite{hafner2022director}, or multi-scale latent predictors in a shared~\cite{zhang2026hierarchical} or level-specific~\cite{zhang2026hjepa} embedding space. Gradients through long unrolled models can have high variance~\cite{parmas2018pipps}, and gradient-based planners can stall in local optima~\cite{bharadhwaj2020model}; differentiable MPC~\cite{amos2018differentiable} instead learns costs and dynamics through the optimizer. H-JEPA~\cite{zhang2026hjepa} runs the same online solver at every level; we instead amortize every level and tailor each proposer to the asymmetry between long-horizon top-level composition and two-step subordinate bridging.

\paragraph{Amortized optimization and offline policy regularization.}
Amortized optimization learns a map from problem instances to solutions~\cite{amos2023tutorial}, in the spirit of reusing computation across inference problems~\cite{gershman2014amortized} and amortized variational inference~\cite{kingma2014auto}. Unlike model-based actor-critic frameworks such as DreamerV2~\cite{hafner2020mastering}, Director~\cite{hafner2022director}, and TD-MPC/TD-MPC2~\cite{hansen2022temporal,hansen2024tdmpc2}, which jointly train reward-maximizing policies alongside dynamics models (TD-MPC and TD-MPC2 still run MPPI at test time), we operate \emph{post-hoc} on a frozen, reward-free H-JEPA. Because optimizing against a learned dynamics model can exploit its errors~\cite{janner2019trust}, we follow behavior-regularized offline RL~\cite{fujimoto2021minimalist} and pair rollout-cost minimization with demonstration regularization.

\paragraph{Multi-choice learning for multi-modal action proposal.}
Goal-conditioned planning is inherently multi-modal: distinct trajectories around obstacles can achieve comparable terminal costs, causing single-head regression to average incompatible modes~\cite{bishop1994mixture}. Instead of iterative denoising as in diffusion policies~\cite{chi2023diffusion}, we follow Multiple Choice Learning (MCL)~\cite{guzman2012multiple,lee2016stochastic,ruppert2017learning} and train $K$ heads with a relaxed winner-take-all (WTA) loss.

\paragraph{Horizon asymmetry and closed-loop sequence modeling.}
ACT~\cite{zhao2023learning} predicts action chunks in parallel and fuses overlapping chunks by temporal ensembling; Decision Transformer~\cite{chen2021decision} predicts actions from return-to-go and past states, and Trajectory Transformer~\cite{janner2021offline} models discretized state-action-reward sequences and plans by beam search. In H-JEPA, subordinate levels bridge two steps, where parallel prediction suffices, whereas the top level plans up to $12$ macro-actions. Unlike these models, our top-level causal Transformer queries a separately pretrained, frozen continuous JEPA predictor \emph{inside} its autoregressive loop, conditioning each macro-action on the evolving imagined trajectory with zero extra rollout overhead.

\section{Preliminaries: Hierarchical JEPA planning}
\label{sec:prelim}

An $L$-level H-JEPA~\cite{zhang2026hjepa} consists of level-specific encoders $E_l$ and single-step Markovian latent predictors $P_l$ ($l \in \{1, \dots, L\}$) operating at doubling temporal strides $\Delta_l = 5 \cdot 2^{l-1}$ environment steps ($\Delta_1 = 5, \dots, \Delta_4 = 40$). Given current and goal observations $(o_0, o_g)$, each level encodes the initial state $z_{l,0} = E_l(o_0)$ and unrolls $P_l$ over a horizon of $H_l$ steps via $\hat{z}_{l,t+1} = P_l(\hat{z}_{l,t}, a_{l,t})$ from $\hat{z}_{l,0} = z_{l,0}$. At the bottom level ($l = 1$), each action $a_{1,t} \equiv u_t \in \mathbb{R}^{5 \times d_a}$ stacks $5$ consecutive raw environment actions; at higher levels ($l \ge 2$), each action is a learned $4$-D macro-action $a_{l,t} \equiv m_{l,t} \in \mathbb{R}^4$ that a frozen action encoder $A_l$, trained jointly with $E_l$ and $P_l$, pools from the two level-$(l{-}1)$ actions it spans. Planning proceeds top-down across two distinct regimes (\cref{fig:arch}a,c):
\begin{itemize}
    \item \textbf{Top level ($l = L$):} Plans a multi-step sequence of $H_L \in [2, 12]$ macro-actions $m_{L,0:H_L-1}$ toward the encoded episode goal $g_L = z_{L,g} = E_L(o_g)$, emitting its first macro-action $\hat{m}_{L,0}$ and immediate subgoal $g_{L-1} = \hat{z}_{L,1}$ to level $L-1$.
    \item \textbf{Subordinate levels ($1 \le l < L$):} Because adjacent strides double ($\Delta_{l+1}/\Delta_l = 2$), each parent macro-action $\hat{m}_{l+1,0}$ decomposes into a short local bridge of $H_l = 2$ child steps tracking the parent subgoal $g_l = \hat{z}_{l+1,1}$, a level-$(l{+}1)$ latent, down to $l = 1$, whose first two action blocks $u_0, u_1$ ($10$ environment steps) are executed before replanning.
\end{itemize}

\paragraph{Mechanics of online gradient-based planning.}
At inference time, standard H-JEPA solves each level $l$ in a nested receding-horizon loop by sampling $N_l \in [25, 800]$ initial action sequences and running $30\text{--}90$ AdamW iterations with early stopping to minimize the terminal squared-$\ell_2$ latent cost:
\begin{equation}
\begin{aligned}
    \mathcal{C}_l\bigl(a_{l, 0:H_l-1}; z_{l,0}, g_l\bigr) &= \bigl\|\Pi_l\bigl(\hat{z}_{l, H_l}\bigr) - g_l\bigr\|_2^2 \\
    &\quad + \lambda_a \bigl\|A_{l+1}(a_{l, 0:H_l-1}) - \hat{m}_{l+1,0}\bigr\|_2^2,
\end{aligned}
    \label{eq:wm_cost}
\end{equation}
where $\Pi_l = E_{l+1}$ maps level-$l$ predictions into the parent's latent space ($\Pi_L = \mathrm{id}$) and $\lambda_a = 0.2$ only at level 1 on Push-T and Cube ($0$ otherwise).
Minimizing \cref{eq:wm_cost} online creates three algorithmic bottlenecks: (i)~\emph{sequential backward unrolling}, as every iteration backpropagates action gradients $\nabla_{a_{l,0:H_l-1}} \mathcal{C}_l$ through $H_l$ unrolled steps of $P_l$ across nested hierarchy calls; (ii)~\emph{non-convex local traps and OOD exploitation}, as local gradient updates from random initializations frequently stall in poor local minima of $\mathcal{C}_l$ or drift into out-of-distribution regions of $P_l$; and (iii)~\emph{top-down subgoal corruption}, as an optimization failure at level $l+1$ can corrupt the subgoal $g_l = \hat{z}_{l+1,1}$ and misdirect all lower levels $1, \dots, l$. \method{} addresses (i) by replacing iterations with one proposal pass, and (ii) by best-of-$K$ selection over behavior-regularized proposals.

\section{Method: \method{}}
\label{sec:method}

We keep all pretrained H-JEPA encoders $\{E_l\}_{l=1}^L$ and predictors $\{P_l\}_{l=1}^L$ strictly frozen.
\Cref{sec:action_planning_model} formulates the amortized planning model shared by all instantiations (learned proposers generate candidates, the frozen world model only scores them); \cref{sec:asymmetry} introduces the homogeneous all-MLP instantiation (\mm{}) and the structural horizon asymmetry between subordinate bridging ($l<L$) and top-level composition ($l=L$); and \cref{sec:tf_ar_wm} details the closed-loop world-model top planner $T_{\phi_L}$ used by \tm{} and \ttarch{}.

\subsection{Amortized action planning model}
\label{sec:action_planning_model}

\paragraph{Best-of-$K$ proposal and one-shot verification.}
To eliminate iterative gradient updates at test time, we introduce at each level $l$ a learned proposer $f_{\phi_l}(c_l)$ mapping the conditioning context $c_l$---$(z_{l,0}, g_l)$ plus the active horizon $h_L \in [2, H_L]$ at $l=L$ or parent context $(z_{l+1,0}, \hat{m}_{l+1,0})$ at $l<L$---to $K_l$ candidate trajectories $\{\hat{a}_{l, 0:H_l-1}^{(k)}\}_{k=1}^{K_l}$ ($K_L=32$; $K_{l<L}=16$).
At inference time, we augment these proposals with a zero-action anchor ($l<L$) or a shifted previous-plan warm-start when available ($l=L$), evaluate the pool in a single batched forward rollout of $P_l$ (or inside the generation loop for $T_{\phi_L}$; \cref{sec:tf_ar_wm}), and select the lowest-cost plan with zero gradient iterations:
\begin{equation}
    k_l^\star = \arg\min_{k} \mathcal{C}_l\bigl(\hat{a}_{l, 0:H_l-1}^{(k)};\; z_{l,0}, g_l\bigr).
    \label{eq:best_of_k}
\end{equation}

\paragraph{Rollout cost and behavior cloning.}
Training $f_{\phi_l}$ solely by imitating dataset actions (behavior cloning, BC) ignores the frozen world model's multi-step transition dynamics, whereas minimizing the frozen rollout cost $\mathcal{C}_l$ alone allows the proposer to exploit out-of-distribution adversarial regions of $P_l$.
Furthermore, because goal-reaching trajectories are inherently multi-modal (\eg, passing left or right around an obstacle), single-head regression averages incompatible action modes.
Given offline training windows $(c_l, a_l^{\text{gt}})$ with goals $g_l$ (at $l = L$ an encoded future frame; at $l < L$ the parent prediction $P_{l+1}(z_{l+1,0}, \hat{m}_{l+1,0})$) and reference actions $a_l^{\text{gt}}$ (macro-actions for $l \ge 2$ and $\hat{m}_{l+1,0}$ pooled by H-JEPA's frozen action encoders $A_l$), we train all $K_l$ heads jointly using a relaxed winner-take-all (WTA) world-model rollout cost plus best-of-$K$ behavior cloning:
\begin{equation}
\begin{aligned}
    \mathcal{L}_l(\phi_l) &= \mathcal{L}_{\text{WTA}} + \lambda_{\text{bc}}\,\mathcal{L}_{\text{BC}}, \\
    \mathcal{L}_{\text{WTA}} &= (1 - \varepsilon)\,\mathcal{C}_l^{(k_l^\star)} + \frac{\varepsilon}{K_l}\sum_{k=1}^{K_l} \mathcal{C}_l^{(k)}, \\
    \mathcal{L}_{\text{BC}} &= \min_{k \in \{1,\dots,K_l\}} \bigl\|\hat{a}_{l, 0:H_l-1}^{(k)} - a_l^{\text{gt}}\bigr\|_2^2,
\end{aligned}
\label{eq:loss}
\end{equation}
where $\mathcal{C}_l^{(k)} = \mathcal{C}_l(\hat{a}_{l,0:H_l-1}^{(k)}; z_{l,0}, g_l)$, $k_l^\star = \arg\min_k \mathcal{C}_l^{(k)}$, $\varepsilon = 0.05$ routes $95\%$ of the world-model gradient to the lowest-cost head while preventing dead heads, and $\lambda_{\text{bc}}$ ($0.05$ at $l=L$, $1$ at $l<L$) regularizes proposals onto the feasible action manifold (trained for 30k AdamW steps on the frozen model's training split; App.~Tab.~S10).

\paragraph{Action-only backward pruning and episode-tail fallback.}
Both during differentiable rollout training (\cref{eq:loss}) and during gradient-based solves, we disable parameter gradient tracking on frozen predictor weights $\theta_{P_l}$, propagating Jacobians strictly along the action pathway $\partial \mathcal{C}_l / \partial a_l$. This eliminates redundant $\nabla_{\theta_{P_l}}\mathcal{C}_l$ accumulation at every unrolled step while yielding bit-identical costs, action gradients, and trajectories ($\max|\Delta a| = 0$; used by all Online GD baselines).
At test time, our proposers replace every top-level solve and every parent-conditioned subordinate solve. Across an episode's replanning calls, H-JEPA decays the active top horizon $h_L$ from $H_L$ to one; once it reaches one, H-JEPA skips the upper levels and issues unconditioned tail \emph{bypass} solves toward the goal (no $\hat{m}_{l+1,0}$; lower horizons then decay from their configured values), for which we retain a 10-iteration GD fallback (9--33\% of subordinate solves, or 75\% on single-top-solve Cube; included in all reported timings).

\subsection{Horizon asymmetry: from MM to TM}
\label{sec:asymmetry}

\paragraph{Homogeneous parallel MLP (\mm{}).}
Given the amortized planning objective in \cref{eq:best_of_k,eq:loss}, the most direct parameterization of $f_{\phi_l}$ is a feed-forward multi-head residual MLP---an input linear projection, $4$ pre-LayerNorm residual blocks (each with two $1024$-dim linear layers and GELU), and a linear output head ($10$ linear layers in total)---that predicts all $H_l$ steps of all $K_l$ candidate trajectories simultaneously in a single parallel pass.
Deploying this parallel MLP across all levels defines the all-MLP instantiation \textbf{\mm{} / \mmm{} / \mmmm{}}:
\begin{itemize}
    \item At \textbf{subordinate levels ($l < L$)}, our proposers map $c_l = (z_{l,0}, z_{l+1,0}, g_l, \hat{m}_{l+1,0})$---where $z_{l+1,0} = E_{l+1}(o_0)$---to $K_l=16$ candidate sequences $\{\hat{a}_{l,0:H_l-1}^{(k)}\}_{k=1}^{16}$.
    \item At the \textbf{top level ($l = L$)}, our MLP proposer maps $[\operatorname{LN}(z_{L,0}), \operatorname{LN}(z_{L,g}), \operatorname{LN}(z_{L,g}-z_{L,0})]$ and the one-hot active horizon indicator $h_L \in [2, H_L]$ to $K_L=32$ macro-action sequences $\{\hat{m}_{L,0:h_L-1}^{(k)}\}_{k=1}^{32}$ in one shot.
\end{itemize}

\paragraph{Horizon asymmetry across levels.}
Analyzing this homogeneous parameterization across levels exposes a fundamental asymmetry in conditioning structure:
\begin{enumerate}
    \item \textbf{Subordinate levels ($l < L$) are boundary-constrained local bridges:} Because $\Delta_{l+1}/\Delta_l = 2$, each subordinate solve is anchored at both ends by the start state $z_{l,0}$, the immediate parent subgoal $g_l = \hat{z}_{l+1,1}$, and the guiding macro-action $\hat{m}_{l+1,0}$, spanning just $H_l = 2$ child steps. Under such tight two-point boundary constraints, a one-shot parallel mapping $c_l \mapsto \hat{a}_{l,0:H_l-1}^{(1:K_l)}$ is well-conditioned and expressive, whereas sequential autoregression (as in \ttarch{} / \ttt{} / \tttt{}) brings no accuracy gain (\cref{sec:main_results,sec:ablations}).
    \item \textbf{Top level ($l = L$) requires state-dependent multi-step composition (up to $H_L = 12$):} In contrast, the top level has no intermediate guide and must compose a long macro-action sequence ($H_2=8, 9, 12$ on L2 Push-T, AntMaze, and FourRoom; $H_L=2$--$3$ only on short-horizon Cube and L4 splits) toward a distant episode goal $z_{L,g}$. Predicting all $H_L$ steps in parallel with an MLP (\mm{}) treats macro-actions $\hat{m}_{L,0:H_L-1}^{(k)}$ as \emph{conditionally independent given only the static endpoint pair $(z_{L,0}, z_{L,g})$}: step $t$ is emitted blind to earlier actions $\hat{m}_{L,0:t-1}^{(k)}$ and the intermediate latent states $\hat{z}_{L,1:t}^{(k)}$ they induce through $P_L$. Over long horizons, navigating obstacles and contact geometries requires conditioning each macro-action on the evolving trajectory.
\end{enumerate}

\subsection{Closed-loop world-model top planner}
\label{sec:tf_ar_wm}

To condition every top-level step on the evolving trajectory while retaining fast parallel MLPs below ($l < L$), our default asymmetric instantiation \textbf{\tm{} (L2) / \tmm{} (L3) / \tmmm{} (L4)} parameterizes the top-level proposer as a 1-layer causal Transformer decoder $T_{\phi_L}$ (\cref{fig:arch}a; denoted \texttt{tf\_ar\_wm} in our ablations), whereas the uniform \textbf{\ttarch{} / \ttt{} / \tttt{}} instantiation deploys the same causal Transformer at every level (conditioned on $(z_{l+1,0}, \hat{m}_{l+1,0})$ instead of $h_L$ for $l < L$).
At the top level $L$, $T_{\phi_L}$ generates $K_L=32$ candidate macro-action sequences autoregressively over steps $t = 0, \dots, h_L-1$ (written below at full horizon $H_L$ for simplicity) with the frozen predictor $P_L$ in the loop, and is trained end-to-end with the same joint objective (\cref{eq:loss}).

\paragraph{World-model-in-the-loop query construction.}
For each candidate mode $k \in \{1,\dots,K_L\}$, we maintain a running imagined latent state $\hat{z}_{L,t}^{(k)}$ (initialized at $\hat{z}_{L,0}^{(k)} = z_{L,0}$) and previous macro-action $\hat{m}_{L,t-1}^{(k)}$ (initialized at $\mathbf{0} \in \mathbb{R}^4$ for $t=0$).
At step $t$, the query token $q_t^{(k)} \in \mathbb{R}^{384}$ encodes the current imagined state and the previous action, while the start state, the goal and the horizon enter through context tokens $M^{(k)}$:
\begin{equation}
\begin{aligned}
    q_t^{(k)} &= \psi_z(\hat{z}_{L,t}^{(k)}) + W_m\,\hat{m}_{L,t-1}^{(k)} \\
    &\quad + e_{\text{step}}[t] + W_\omega\,\omega_{\text{cand}}[k], \\
    M^{(k)} &= \bigl[\psi_0(z_{L,0}),\, \psi_g(z_{L,g}),\, \psi_\Delta(z_{L,g} - z_{L,0}), \\
    &\qquad W_h \operatorname{onehot}(h_L),\, \omega_{\text{cand}}[k]\bigr],
\end{aligned}
\label{eq:tf_query}
\end{equation}
where each $\psi_{\cdot}$ is a LayerNorm followed by a linear projection to $\mathbb{R}^{384}$, $W_m$, $W_h$ and $W_\omega$ are linear projections, $e_{\text{step}}[t]$ sums learned embeddings of the step index and of the number of remaining steps, $\omega_{\text{cand}}[k] \in \mathbb{R}^{384}$ is a learned candidate-mode embedding, and the context tokens additionally carry learned token-type embeddings.
Note that clamping $\hat{z}_{L,t}^{(k)} \equiv z_{L,0}$ and $\hat{m}_{L,t-1}^{(k)} \equiv \mathbf{0}$ reduces \cref{eq:tf_query} to a parallel self-attention baseline (\texttt{tf\_parallel}), whereas feeding $\hat{m}_{L,t-1}^{(k)}$ while keeping $\hat{z}_{L,t}^{(k)} \equiv z_{L,0}$ yields open-loop action autoregression (\texttt{tf\_ar\_action}).

\begin{table*}[t]
\centering
\caption{\textbf{Main results on 30 frozen H-JEPA models.} Success rate (\%, mean $\pm$ s.e.\ over three seeds; 250 tasks per seed on Push-T, 50 elsewhere) and time per episode (s; seed 42, 10 tasks) of Online GD and the three \method{} instantiations (MM, TT, TM: e.g.\ \mmm{}, \ttt{}, \tmm{} at L3). Bold: best success per column. Speed-up: range over the three instantiations. Per-seed results, paired tests and all timings: App.~Tab.~S2, S4 and S7.}
\label{tab:main}
\small
\setlength{\tabcolsep}{2.6pt}

\begin{tabularx}{\textwidth}{l@{\hspace{5pt}}YYYYYYYYYY}
\toprule
 & \multicolumn{4}{c}{Two levels (L2)} & \multicolumn{3}{c}{Three levels (L3)} & \multicolumn{3}{c}{Four levels (L4)} \\
\cmidrule(lr){2-5}\cmidrule(lr){6-8}\cmidrule(l){9-11}
 & Push-T & FourRoom & AntMaze & Cube & Push-T & FourRoom & AntMaze & Push-T & FourRoom & AntMaze \\
{\footnotesize\color{black!60}top horizon $H_L$} & \leavevmode{\footnotesize\color{black!60}8} & \leavevmode{\footnotesize\color{black!60}12} & \leavevmode{\footnotesize\color{black!60}9} & \leavevmode{\footnotesize\color{black!60}2} & \leavevmode{\footnotesize\color{black!60}4} & \leavevmode{\footnotesize\color{black!60}6} & \leavevmode{\footnotesize\color{black!60}5} & \leavevmode{\footnotesize\color{black!60}2} & \leavevmode{\footnotesize\color{black!60}3} & \leavevmode{\footnotesize\color{black!60}3} \\
\midrule
\multicolumn{11}{l}{\textit{Success rate (\%)} $\uparrow$} \\
Online GD~\cite{zhang2026hjepa} & 40.8\se{2.1} & 83.3\se{4.1} & 45.3\se{2.9} & 48.0\se{4.2} & 15.2\se{1.1} & 92.0\se{3.1} & 58.7\se{2.9} & 0.1\se{0.1} & 91.3\se{2.9} & 62.0\se{4.2} \\
\multicolumn{11}{l}{\method{} (ours) \footnotesize\color{black!60}(top / subordinate proposers; M: MLP, T: Transformer)} \\
\rowcolor{oursrow}\quad MM & 55.9\se{1.9} & 92.0\se{1.2} & \textbf{78.7}\se{3.3} & \textbf{80.7}\se{2.9} & 16.4\se{2.8} & 99.3\se{0.7} & \textbf{80.7}\se{5.7} & 0.5\se{0.5} & \textbf{94.0}\se{2.0} & 83.3\se{4.7} \\
\rowcolor{oursrow}\quad TT & 63.2\se{1.7} & \textbf{100.0}\se{0.0} & 78.0\se{8.0} & 75.3\se{2.4} & \textbf{18.3}\se{1.0} & \textbf{100.0}\se{0.0} & \textbf{80.7}\se{2.7} & 0.0\se{0.0} & 86.7\se{5.3} & \textbf{88.7}\se{0.7} \\
\rowcolor{oursrow}\quad TM & \textbf{64.7}\se{1.9} & \textbf{100.0}\se{0.0} & \textbf{78.7}\se{3.5} & 74.0\se{2.3} & 17.9\se{1.5} & \textbf{100.0}\se{0.0} & \textbf{80.7}\se{4.8} & \textbf{1.2}\se{1.2} & 92.0\se{1.2} & 84.0\se{4.2} \\
\midrule
\multicolumn{11}{l}{\textit{Time per episode (s)} $\downarrow$} \\
Online GD & 25.3 & 45.1 & 78.9 & 3.4 & 20.3 & 36.6 & 84.3 & 18.8 & 46.2 & 92.5 \\
\rowcolor{oursrow}\quad MM & 1.43 & 2.76 & 5.23 & 1.18 & 1.49 & 3.02 & 6.09 & 1.91 & 4.21 & 7.12 \\
\rowcolor{oursrow}\quad TT & 1.37 & 2.93 & 5.70 & 1.16 & 1.49 & 3.19 & 6.28 & 1.96 & 4.49 & 7.30 \\
\rowcolor{oursrow}\quad TM & 1.34 & 2.93 & 5.69 & 1.19 & 1.48 & 3.10 & 6.23 & 1.91 & 4.29 & 7.15 \\
Speed-up ($\times$) & {\footnotesize\bfseries 17.7--18.8} & {\footnotesize\bfseries 15.4--16.4} & {\footnotesize\bfseries 13.8--15.1} & {\footnotesize\bfseries 2.9--3.0} & {\footnotesize\bfseries 13.6--13.7} & {\footnotesize\bfseries 11.5--12.1} & {\footnotesize\bfseries 13.4--13.8} & {\footnotesize\bfseries 9.6--9.9} & {\footnotesize\bfseries 10.3--11.0} & {\footnotesize\bfseries 12.7--13.0} \\
\bottomrule
\end{tabularx}

\vspace{-6pt}
\end{table*}%
\paragraph{1-layer causal decoder and zero-extra-rollout selection.}
The query sequence $Q_{0:t}^{(k)} = [q_0^{(k)}, \dots, q_t^{(k)}]$ passes through one standard pre-LayerNorm causal Transformer decoder layer~\cite{vaswani2017attention,xiong2020layer} ($d_{\text{model}}=384$, $n_{\text{head}}=6$, $d_{\text{ffn}}=1536$, GELU) that cross-attends to $M^{(k)}$; LayerNorm and a 2-layer GELU MLP head map its step-$t$ output to $\hat{m}_{L,t}^{(k)}$, squashed into the macro-action box by a sigmoid.
Immediately after emitting $\hat{m}_{L,t}^{(k)}$, before constructing the query $q_{t+1}^{(k)}$, $T_{\phi_L}$ queries the frozen level-$L$ JEPA predictor $P_L$ for a single step (advancing any warm-start plan in the same batch):
\begin{equation}
    \hat{z}_{L, t+1}^{(k)} = P_L\bigl(\hat{z}_{L, t}^{(k)},\; \hat{m}_{L, t}^{(k)}\bigr).
    \label{eq:advance_jepa}
\end{equation}
Because H-JEPA's higher-level predictors use single-step Markovian transitions (\cref{sec:prelim}), chaining \cref{eq:advance_jepa} step-by-step is numerically bit-identical to a full multi-step batch rollout ($\max |\Delta \hat{z}| = 0.0$).
Consequently, when step $t = H_L - 1$ completes, the entire imagined trajectory $\hat{z}_{L, 1:H_L}^{(k)}$ has \emph{already been computed inside the generation loop}.
Selecting the best candidate $k_L^\star = \arg\min_k \mathcal{C}_L^{(k)}$ and emitting $\hat{m}_{L,0}^{(k_L^\star)}$ and $g_{L-1} = \hat{z}_{L,1}^{(k_L^\star)}$ to level $L-1$ therefore requires zero additional world-model rollouts: selection costs $0.13$\,ms instead of a separate batched rollout ($30$\,ms for $H_L=8$ on Push-T; \cref{tab:ablation}).
In our ablations (\cref{sec:ablations}), we compare this 1-layer decoder ($3.4\text{--}3.7$M parameters across tasks) against a 4-layer variant parameter-matched to the 4-block MLP ($10.1$M parameters), confirming that a single causal layer matches 4 layers in success while halving training time.

\section{Experiments}
\label{sec:experiments}

We evaluate \method{} on (i) multi-depth planning speed and task success across 2-, 3-, and 4-level hierarchies (\cref{sec:main_results}), (ii) architecture and single-call latency ablations (\cref{sec:ablations}), (iii) mechanism studies on OOD exploitation, candidate budget $K$, and GD refinement (\cref{sec:ablations}), and (iv) per-level latency across depths (\cref{sec:latency}).

\subsection{Experimental setup}
\label{sec:setup}

\paragraph{World models, planners, and metrics.}
We train H-JEPA~\cite{zhang2026hjepa} with the official configurations across three seeds (42, 43, 44) and keep all encoders $E_l$, predictors $P_l$, and action encoders $A_l$ strictly frozen: two-level (L2) models on Push-T, FourRoom, Visual AntMaze, and OGBench Cube~\cite{park2024ogbench}, and three- and four-level (L3, L4) models on Push-T, FourRoom, and AntMaze (the official Cube L3/L4 splits have unit upper horizons and reduce to L2). This yields 30 frozen models with top horizons $H_L \in [2, 12]$ (\cref{tab:main}; App.~Tab.~S1), evaluated on 50 official tasks (+200 held-out tasks/seed on Push-T).
All planners share the same frozen checkpoints, receding-horizon loop, and tail bypass rule (\cref{sec:action_planning_model}):
\emph{Online GD}~\cite{zhang2026hjepa} runs the official H-JEPA solver with per-environment budgets (App.~Tab.~S1) and exact action-only gradients;
\emph{MM} (\mm{} / \mmm{} / \mmmm{}) deploys the 4-block residual MLP $f_{\phi_l}$ (\cref{sec:asymmetry}) at every level;
\emph{TT} (\ttarch{} / \ttt{} / \tttt{}) deploys the 1-layer closed-loop Transformer (\cref{sec:tf_ar_wm}; denoted \texttt{tf\_ar\_wm} in ablations) at every level;
and \emph{TM} (\tm{} / \tmm{} / \tmmm{}) pairs the top-level closed-loop Transformer $T_{\phi_L}$ with residual MLPs $f_{\phi_l}$ at all subordinate levels ($K_L=32$, $K_{l<L}=16$, trained once per frozen model).
We report success rate as mean $\pm$ s.e.\ over the three seeds, test paired differences via exact two-sided McNemar tests, and time episodes sequentially on one H200 GPU (seed 42, 10 tasks).

\subsection{Multi-depth efficiency and performance}
\label{sec:main_results}

\begin{table}[t]
\centering
\caption{\textbf{Architecture ablations on Push-T L2} (seed 42, 250 paired tasks; $H_2=8$, $H_1=2$). Group rows: top-level proposer, its parameters and the time of one top-level call ($K_2=32$; scoring part in parentheses; breakdown in App.~Tab.~S11). $^{\dagger}p<0.05$, $^{\ddagger}p<0.01$ \vs Online GD (McNemar). All 14 variants: App.~Tab.~S6.}
\label{tab:ablation}
\footnotesize
\setlength{\tabcolsep}{3pt}

\begin{tabularx}{\linewidth}{lRRR}
\toprule
Level-1 proposer & Succ.\ (\%) & $\Delta_{\text{GD}}$ (pp) & Ep.\ (s) \\
\midrule
\multicolumn{4}{l}{\textit{Top:} Online GD} \\
\quad Online GD & 40.0 & ref. & 25.3 \\
\quad MLP (top: 10 GD it.) & 38.8 & $-1.2$ & 5.55 \\
\addlinespace[2pt]
\multicolumn{2}{l}{\textit{Top:} MLP} & \multicolumn{2}{r}{{\color{black!60}10.1M, 30.5\,ms (30.0)}} \\
\quad Online GD & 49.6\rlap{$^{\dagger}$} & $+9.6$ & 6.64 \\
\rowcolor{oursrow}\quad MLP (\mm) & 56.4\rlap{$^{\ddagger}$} & $+16.4$ & 1.43 \\
\quad \texttt{tf\_parallel} (4L) & 57.6\rlap{$^{\ddagger}$} & $+17.6$ & 1.34 \\
\quad \texttt{tf\_ar\_action} (4L) & 56.4\rlap{$^{\ddagger}$} & $+16.4$ & 1.33 \\
\quad \texttt{tf\_ar\_wm} (4L) & 53.6\rlap{$^{\ddagger}$} & $+13.6$ & 1.31 \\
\addlinespace[2pt]
\multicolumn{2}{l}{\textit{Top:} \texttt{tf\_parallel} (4L)} & \multicolumn{2}{r}{{\color{black!60}10.1M, 35.2\,ms (30.1)}} \\
\quad MLP & 64.4\rlap{$^{\ddagger}$} & $+24.4$ & 1.34 \\
\addlinespace[2pt]
\multicolumn{2}{l}{\textit{Top:} \texttt{tf\_ar\_action} (4L)} & \multicolumn{2}{r}{{\color{black!60}10.1M, 56.3\,ms (29.8)}} \\
\quad MLP & 58.4\rlap{$^{\ddagger}$} & $+18.4$ & 1.40 \\
\addlinespace[2pt]
\multicolumn{2}{l}{\textit{Top:} \texttt{tf\_ar\_wm} (4L)} & \multicolumn{2}{r}{{\color{black!60}10.1M, 57.9\,ms (\textbf{0.14})}} \\
\quad MLP & 67.6\rlap{$^{\ddagger}$} & $+27.6$ & 1.39 \\
\quad \texttt{tf\_ar\_wm} (4L) & \textbf{68.0}\rlap{$^{\ddagger}$} & $+28.0$ & 1.43 \\
\addlinespace[2pt]
\multicolumn{2}{l}{\textit{Top:} \texttt{tf\_ar\_wm} (1L)} & \multicolumn{2}{r}{{\color{black!60}3.4M, 41.5\,ms (\textbf{0.13})}} \\
\rowcolor{oursrow}\quad MLP (\tm) & 66.4\rlap{$^{\ddagger}$} & $+26.4$ & 1.34 \\
\rowcolor{oursrow}\quad \texttt{tf\_ar\_wm} (1L, \ttarch) & 65.6\rlap{$^{\ddagger}$} & $+25.6$ & 1.37 \\
\bottomrule
\end{tabularx}

\vspace{-6pt}
\end{table}

\begin{figure}[t]
\centering
\includegraphics[width=\columnwidth]{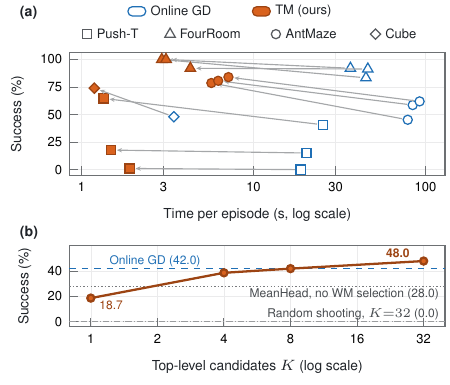}
\caption{\textbf{(a) Speed and success} of Online GD (hollow) and \method{}-TM (filled) on all ten settings of \cref{tab:main} (mean over three seeds; time: seed 42). \textbf{(b) Best-of-$K$ selection} over the \mm{} top-level proposals on Push-T L2 (150 tasks; \cref{tab:mechanism}): success grows with $K$, matches Online GD at $K=8$ and exceeds averaging the proposals (\emph{MeanHead}) and random shooting.}
\label{fig:pareto}
\vspace{-6pt}
\end{figure}

\paragraph{End-to-end speed-up.}
Across the nine long-horizon settings in \cref{tab:main}, \method{} is $9.6$--$18.8\times$ faster than Online GD, cutting episode runtime from $18.8$--$92.5$\,s to $1.3$--$7.3$\,s (\cref{tab:main,fig:pareto}a). On Cube, whose episodes require a single two-step top-level solve ($H_2=2$), the speed-up is $2.9$--$3.0\times$. Among the long-horizon settings, it is largest on Push-T and FourRoom L2 ($17.7$--$18.8\times$ and $15.4$--$16.4\times$) and smallest on Push-T L4 ($9.6$--$9.9\times$). MM, TT, and TM execute within $9\%$ of each other in wall-clock time: one top-level call takes $30.5$\,ms with the MLP and $41.5$\,ms with the 1-layer Transformer, whose in-loop rollout replaces the separate scoring rollout (\cref{tab:ablation}).

\paragraph{Task success and escaping local minima.}
With all encoders and predictors frozen, success gains stem purely from planning optimization. No \method{} instantiation is significantly worse than Online GD on any setting (\cref{tab:main}; largest drop: TT on FourRoom L4, $-4.7$\,pp, $p=0.26$, App.~Tab.~S4). Excluding Push-T L4, where H-JEPA's 40-step stride ($\Delta_4{=}40$) aliases contact dynamics so all planners fail ($\le 1.2\%$), all three instantiations significantly outperform Online GD on the same seven of the nine remaining settings, gaining up to $+23.9$\,pp on Push-T L2 ($64.7\%$ \vs $40.8\%$ for TM), $+16.7$\,pp on FourRoom L2 ($100.0\%$ \vs $83.3\%$), $+32.7$\,pp on Cube L2 ($80.7\%$ \vs $48.0\%$ for MM), and $+33.3$\,pp on AntMaze L2 ($78.7\%$ \vs $45.3\%$ for TM, $p=1.8\times10^{-10}$). These gains extend to deeper hierarchies ($+22.0$\,pp on AntMaze L3 and L4), while matching Online GD on Push-T L3 (TM: $17.9\%$ \vs $15.2\%$, $p=0.13$) and FourRoom L4 ($92.0\%$ \vs $91.3\%$; visual rollouts in App.~Fig.~S1). On Push-T L2 the gain holds on both task sets: TM reaches $64.0\%$ on the official and $64.8\%$ on the held-out tasks, against $40.7\%$ and $40.8\%$ for Online GD (App.~Tab.~S3).

\paragraph{Horizon asymmetry across tasks and depths.}
Comparing MM, TT, and TM across \cref{tab:main} validates the horizon asymmetry analyzed in \cref{sec:asymmetry}. On long top-level horizons, closed-loop rollout conditioning in TM significantly outperforms one-shot open-loop prediction in MM, gaining $+8.8$\,pp on Push-T L2 ($p=1.7\times10^{-5}$; $H_L=8$) and $+8.0$\,pp on FourRoom L2 ($p=4.9\times10^{-4}$; $H_L=12$), with TT achieving similar gains ($+7.3$ and $+8.0$\,pp). Conversely, when $H_L \in \{2, 3\}$---on Cube ($H_2=2$) or on 4-level hierarchies where the top level spans only $2$--$3$ macro-steps---top-level planning reduces to short-horizon bridging where one-shot prediction is already well-constrained: MM reaches $80.7\%$ on Cube ($p=0.14$ \vs TM) and $94.0\%$ on FourRoom L4 ($p=0.55$); MM also matches TM on the remaining L3 and AntMaze settings. Extending the Transformer to subordinate levels (TT) brings no significant gain over TM on any setting (and is significantly worse on Push-T L4, $-1.2$\,pp, $p=0.004$; App.~Tab.~S4), supporting TM as our default.

\subsection{Controlled ablations and mechanism analysis}
\label{sec:ablations}

\paragraph{Full-hierarchy amortization.}
On 2-level Push-T (seed 42, 250 paired tasks; Online GD and MLP rows of \cref{tab:ablation}),
amortizing only Level 1 leaves top-level GD as the bottleneck ($5.55$\,s, $38.8\%$ \vs $40.0\%$); across the four L2 benchmarks it is $2.5$--$4.6\times$ faster and changes success by $-7.3$ to $+27.3$\,pp (App.~Tab.~S5). Amortizing only the top level raises success to $49.6\%$ ($+9.6$\,pp, $p<0.05$) but leaves Level-1 gradient descent at $6.64$\,s. Jointly amortizing both levels (\mm{}) is $17.7\times$ faster than Online GD ($1.43$\,s) and $+16.4$\,pp more successful ($56.4\%$, $p=5.5\times10^{-5}$).

\paragraph{WARP-JEPA design space: parallel (\mm) \vs in-loop rollout (\tm).}
With an MLP at level 1, all learned top-level proposers in \cref{tab:ablation} substantially outperform Online GD ($+16.4$ to $+27.6$\,pp). Our parallel instantiation \mm{} generates all $K_2=32$ candidates in $0.6$\,ms ($30.5$\,ms with post-hoc rollout scoring; App.~Tab.~S11; $+16.4$\,pp over GD). For long horizons ($H_2=8$), adding parallel step self-attention (\texttt{tf\_parallel}, $+24.4$\,pp over GD) helps more than open-loop action autoregression (\texttt{tf\_ar\_action}, $+18.4$\,pp), while feeding back imagined world-model states $\hat{z}_{2,t}$ inside the loop (\texttt{tf\_ar\_wm}, 4L) reaches $67.6\%$ ($+27.6$\,pp over GD; $+11.2$\,pp over \mm{}, $p=0.002$). Reusing its in-loop rollout cuts candidate scoring from $\sim30.0$ to $0.13$--$0.14$\,ms, and our 1-layer default \tm{} ($3.4$M) lowers call latency ($57.9 \to 41.5$\,ms) and halves training time ($2.1$ \vs $4.2$\,h; App.~Tab.~S10) while retaining $+26.4$\,pp over GD ($66.4\%$). At Level 1 ($H_1=2$; MLP top-level rows of \cref{tab:ablation}), 4-layer Transformers ($53.6\%$--$57.6\%$, $p \ge 0.42$ \vs \mm{}) bring no gain over the parallel MLP. Under the 1-layer top-level Transformer, the same holds: a level-1 Transformer (\ttarch{}, $65.6\%$, $1.37$\,s) matches the level-1 MLP (\tm{}, $66.4\%$, $1.34$\,s), so the cheaper subordinate MLP costs no success.

\begin{table}[t]
\centering
\caption{\textbf{Mechanism analysis on Push-T L2} (released checkpoint; level 1: MLP, except for Online GD). Offline: median terminal cost $\mathcal{C}_2$ and first-action MSE on 192 validation segments. Closed loop: 150 tasks, RTX 4090; $^{\ddagger}p<0.01$ \vs \mm{}. $^*$Reversed on 200 new tasks ($49.5\%$ \vs $53.5\%$). All 18 variants: App.~Tab.~S12.}
\label{tab:mechanism}
\footnotesize
\setlength{\tabcolsep}{3.0pt}

\begin{tabularx}{\linewidth}{Xrrrr}
\toprule
 & \multicolumn{2}{c}{Offline} & \multicolumn{2}{c}{Closed loop} \\
\cmidrule(lr){2-3}\cmidrule(l){4-5}
Top-level solver & $\mathcal{C}_2$ & MSE & Succ.\ (\%) & Time (s) \\
\midrule
Expert actions & 215.0 & 0.00 & -- & -- \\
Online GD, 30--90 it. & 16.9 & 4.23 & 42.0 & 44.3 \\
Random shooting, $K{=}32$ & 275.0 & 6.74 & 0.0\rlap{$^{\ddagger}$} & 2.33 \\
\addlinespace[2pt]
\multicolumn{5}{l}{\textit{MLP proposals} ($\lambda_{\mathrm{bc}}{=}0.05$)} \\
\quad Mean of 32 (MeanHead) & 314.8 & 0.97 & 28.0\rlap{$^{\ddagger}$} & 2.21 \\
\quad Best of $K{=}1$ & 278.9 & 1.20 & 18.7\rlap{$^{\ddagger}$} & 2.25 \\
\quad Best of $K{=}4$ & 165.3 & 0.91 & 38.7 & 2.28 \\
\quad Best of $K{=}8$ & 130.2 & 0.89 & 42.0 & 2.34 \\
\rowcolor{oursrow}\quad Best of $K{=}32$ (\mm) & 73.7 & 1.25 & 48.0 & 2.31 \\
\addlinespace[2pt]
\multicolumn{5}{l}{\textit{Variants of} \mm{}} \\
\quad $\lambda_{\mathrm{bc}}{=}0$ & 74.7 & 3.14 & 42.7 & 2.26 \\
\quad $+5$ GD steps & 23.2 & 1.17 & 52.0\rlap{$^*$} & 5.96 \\
\bottomrule
\end{tabularx}

\vspace{-6pt}
\end{table}

\paragraph{OOD exploitation, best-of-$K$ selection, and zero-gradient sufficiency.}
\Cref{tab:mechanism} and \cref{fig:pareto}b isolate why proposal selection outperforms online gradient optimization on the released Push-T L2 checkpoint:
(1) \emph{Escaping OOD predictor exploitation}: Online GD drives terminal world-model cost $\mathcal{C}_2$ to $16.9$ ($2.5$ at 90 iterations; App.~Tab.~S12)---$13\times$--$86\times$ below the ground-truth expert trajectory ($\mathcal{C}_2 = 215.0$)---yet its actions deviate strongly from the expert manifold ($\text{MSE} = 4.23$), yielding $42.0\%$ success. Training without behavior regularization ($\lambda_{\mathrm{bc}}=0$) similarly drifts off-manifold ($\text{MSE}=3.14$, $42.7\%$), whereas $\lambda_{\mathrm{bc}}=0.05$ anchors proposals near feasible actions ($\text{MSE}=1.25$) at a $4.4\times$ higher rollout cost ($\mathcal{C}_2=73.7$) and reaches $48.0\%$ success. Although neither difference is significant on 150 tasks, a lower predicted cost thus does not buy higher success, the discrepancy noted in \cref{sec:intro}.
(2) \emph{Best-of-$K$ selection \vs mode averaging and shooting}: Zero-order random shooting from the empirical prior ($K=32$, 0 GD) achieves $0/150$: best-of-$K$ selection fails without a learned proposal distribution. Averaging the 32 proposal heads without world-model selection (\emph{MeanHead}) achieves the lowest Euclidean action MSE ($0.97$) but averages incompatible modes, inflating $\mathcal{C}_2$ to $314.8$ and dropping success to $28.0\%$ ($-20.0$\,pp, $p=2.9\times 10^{-4}$). With world-model selection (\cref{fig:pareto}b), scaling $K$ from $1 \to 4 \to 8 \to 32$ monotonically reduces $\mathcal{C}_2$ ($278.9 \to 73.7$) and lifts success ($18.7\% \to 38.7\% \to 42.0\% \to 48.0\%$).
(3) \emph{Zero-gradient sufficiency ($R=0$)}: Appending $R=2$ or $R=5$ online GD steps to the $K=32$ proposals further lowers $\mathcal{C}_2$ ($73.7 \to 50.2 \to 23.2$; App.~Tab.~S12) at $1.6\times$--$2.6\times$ higher latency, yet brings no closed-loop gain (150 exploratory tasks: $52.0\%$ \vs $48.0\%$, $p=0.44$; 200 confirmatory: $49.5\%$ \vs $53.5\%$, $p=0.38$): once best-of-$K$ selection locates a feasible mode basin, local unregularized GD steps mainly exploit predictor curvature.

\subsection{Hierarchical latency breakdown}
\label{sec:latency}
\begin{table}[t]
\centering
\caption{\textbf{AntMaze time per episode (s) by level} (seed 42, 10 tasks). Middle: levels 2 to $L-1$; Rest: encoding, simulation, overhead. All settings: App.~Tab.~S8.}
\label{tab:latency}
\small
\setlength{\tabcolsep}{3.4pt}

\begin{tabularx}{\linewidth}{llRRRRR}
\toprule
 & & \multicolumn{3}{c}{Solver time per level} & & \\
\cmidrule(lr){3-5}
Depth & Planner & Top & Middle & L1 & Rest & Total \\
\midrule
L2 & Online GD & 47.6 & -- & 28.3 & 2.9 & 78.9 \\
\rowcolor{oursrow} & \textbf{\tm} & 1.38 & -- & 1.39 & 2.92 & 5.69 \\
\addlinespace[2pt]
L3 & Online GD & 32.8 & 19.9 & 28.4 & 3.1 & 84.3 \\
\rowcolor{oursrow} & \textbf{\tmm} & 0.81 & 1.06 & 1.40 & 2.96 & 6.23 \\
\addlinespace[2pt]
L4 & Online GD & 19.3 & 39.7 & 30.4 & 3.0 & 92.5 \\
\rowcolor{oursrow} & \textbf{\tmmm} & 0.50 & 2.60 & 1.09 & 2.97 & 7.15 \\
\bottomrule
\end{tabularx}

\vspace{-6pt}
\end{table}

\Cref{tab:latency} decomposes per-episode runtime on AntMaze by hierarchy level across L2, L3, and L4. Online GD spends $19.3$--$47.6$\,s in per-level solvers ($96.3$--$96.7\%$ of episode time; $96.9$--$97.3\%$ total planner share, App.~Tab.~S7), so amortizing a single level leaves most latency intact. \method{} (\tm{}, \tmm{}, \tmmm{}) reduces solver time at every level to $0.50$--$1.40$\,s ($2.60$\,s across the two middle levels at L4); top-level and level-1 solver times drop from $19.3$--$47.6$ to $0.50$--$1.38$\,s and from $28.3$--$30.4$ to $1.09$--$1.40$\,s. These times include the episode-tail GD bypass solves ($9$--$21\%$ of subordinate solves; App.~Tab.~S9). Roughly $3.0$\,s per episode goes to encoding and simulation outside the solvers ($42$--$51\%$ of \method{} runtime), bounding further speed-ups: at L4, total solver time falls $21\times$, episode time only $12.9\times$. From L2 to L4, Online GD gains $13.6$\,s per episode, mostly in middle-level solvers, but \method{} only $1.46$\,s, so the cost limitation of \cref{sec:intro} is addressed at every depth ($12.9$--$13.9\times$).

\section{Conclusion}
\label{sec:conclusion}

We presented \textbf{\method{}}, an amortized hierarchical planning framework for frozen H-JEPA world models. Across 30 frozen 2-, 3-, and 4-level models, its all-MLP, all-Transformer, and asymmetric instantiations (MM, TT, TM) accelerate episode inference by $9.6$--$18.8\times$ on long-horizon tasks ($2.9$--$3.0\times$ on Cube) and improve task success by up to $+33.3$\,pp over online gradient descent. Guided by the horizon asymmetry between multi-step top-level composition and two-step subordinate bridging, our asymmetric default TM pairs a 1-layer world-model-in-the-loop causal Transformer at the top level (scoring candidates with zero extra rollouts) with parallel residual MLPs below.

\paragraph{Limitations.}
\method{} adds a post-hoc training step after each H-JEPA model is trained (one proposer per level, 0.2--3.6\,h for default proposers, up to 4.2\,h for 4L ablations on one H200; App.~Tab.~S10) and leaves the world model unchanged, so it cannot repair model errors: with H-JEPA's 40-step top stride on Push-T L4, every planner fails ($\le 1.2\%$). End-of-episode bypass solves still run 10 GD iterations (9--33\% of subordinate solves, 75\% of level-1 solves on Cube), and about 3\,s per AntMaze episode is spent outside the solvers.

{
    \small
    \bibliographystyle{ieeenat_fullname}
    \bibliography{main}
}

\clearpage
\appendix
\renewcommand{\thetable}{S\arabic{table}}
\renewcommand{\thefigure}{S\arabic{figure}}
\setcounter{table}{0}\setcounter{figure}{0}
\begin{center}{\Large\bfseries Appendix}\end{center}

This appendix reports the evaluation protocol (\cref{sec:s_protocol}), the complete results behind Tables~1--4 (\cref{sec:s_results,sec:s_ablation}), the timing breakdown (\cref{sec:s_timing}) and proposer details (\cref{sec:s_proposers}). Tables~S1--S12 are referenced from the main text by these numbers.

\section{Evaluation protocol}
\label{sec:s_protocol}

\paragraph{World models.}
All H-JEPA models are trained with the official configurations for model seeds 42, 43 and 44 on single NVIDIA H200 GPUs and are kept frozen afterwards. The two-level (L2) models cover Push-T, FourRoom, Visual AntMaze and OGBench Cube; the three- and four-level (L3, L4) models cover Push-T, FourRoom and AntMaze. \Cref{tab:s_config} lists the planning horizons of every level.

\paragraph{Tasks.}
Every model is evaluated on the 50 official tasks of its benchmark. On Push-T we add 200 held-out tasks per seed (50 for each of the task seeds 1003--1006), generated with the official task generator from the same validation episodes; the generator reproduces the official task file with its default seed. Push-T results therefore use 250 tasks per seed, the other benchmarks 50.

\paragraph{Online GD baseline.}
Online GD is the released H-JEPA planner with the official evaluation configuration of each benchmark (\cref{tab:s_config}): per level, AdamW over the listed number of sampled candidates for at most 90 iterations with early stopping (at least 30 iterations, patience 5), replanning every 10 environment steps. Gradients are taken with respect to the actions only; this is bit-identical to the released planner (costs, action gradients and actions are checked in every run) and faster, so all speed-ups are measured against this improved baseline.

\paragraph{Statistics.}
Success rates are averaged over the three model seeds (standard error $=$ standard deviation$/\sqrt{3}$). Differences between planners are pooled over all (seed, task) pairs and tested with the exact two-sided McNemar test on the discordant pairs.

\paragraph{Timing.}
Time per episode is measured in dedicated sequential runs on one H200 GPU for model seed 42 on the first 10 official tasks, with CUDA-synchronized timers around every planner and encoder call. It includes encoding and simulation. The success evaluations were partly run concurrently on shared GPUs, so their wall-clock times are not used.

\begin{table}[t]
\centering
\caption{\textbf{Hierarchies and Online GD budgets} (official configurations; horizons and candidates from level~1 up). Online GD: AdamW, 30--90 iterations per solve (early stop, patience 5), replanning every 10 steps.}
\label{tab:s_config}
\footnotesize
\setlength{\tabcolsep}{3pt}
\begin{tabular*}{\columnwidth}{@{\extracolsep{\fill}}llcccc@{}}
\toprule
Benchmark & Depth & \shortstack{Horizons\\$H_1,\dots,H_L$} & \shortstack{GD candidates\\per level} & \shortstack{Tasks\\/ seed} & Tasks \\
\midrule
Push-T & L2 & 2, 8 & 50, 400 & 250 & 750 \\
FourRoom & L2 & 2, 12 & 100, 800 & 50 & 150 \\
AntMaze & L2 & 4, 9 & 25, 200 & 50 & 150 \\
Cube & L2 & 2, 2 & 400, 100 & 50 & 150 \\
Push-T & L3 & 2, 2, 4 & 25, 50, 200 & 250 & 750 \\
FourRoom & L3 & 2, 2, 6 & 50, 200, 200 & 50 & 150 \\
AntMaze & L3 & 4, 2, 5 & 100, 100, 400 & 50 & 150 \\
Push-T & L4 & 2, 2, 2, 2 & 200, 800, 800, 800 & 250 & 750 \\
FourRoom & L4 & 2, 2, 2, 3 & 50, 200, 200, 400 & 50 & 150 \\
AntMaze & L4 & 3, 2, 2, 3 & 50, 200, 25, 50 & 50 & 150 \\
\bottomrule
\end{tabular*}
\end{table}

\section{Complete results}
\label{sec:s_results}

\begin{figure*}[t]
\centering
\includegraphics[width=\textwidth]{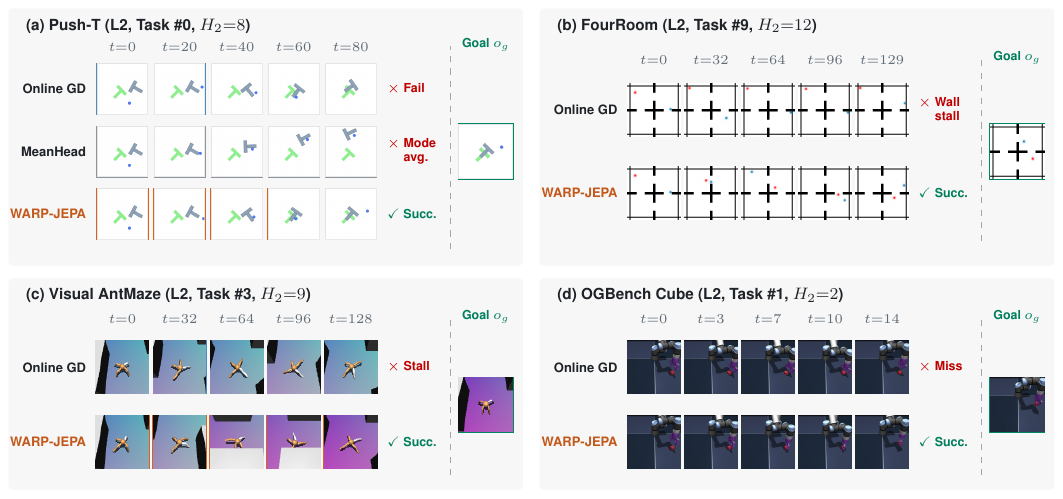}
\caption{\textbf{Qualitative closed-loop visual rollouts across all four benchmarks} (L2, seed 42, \method{} = \mm{}; step-aligned frames from $t=0$ to $T_{\text{succ}}$ alongside target observation $o_g$). \textbf{(a) Push-T (Task \#0):} Online GD stalls with the T-block misaligned, and \emph{MeanHead} (averaging the $K=32$ proposal heads without world-model selection) averages incompatible left/right modes and pushes the block away from the target, whereas \method{} selects a feasible mode and aligns the T-block with $o_g$ at $t=80$. \textbf{(b) FourRoom (Task \#9):} Online GD gets trapped against the first interior wall, while \method{} routes through both doorway subgoals to reach the bottom-right room ($t=129$). \textbf{(c) Visual AntMaze (Task \#3):} Online GD stalls in the initial corridor, whereas \method{} rounds the corner into the purple goal corridor ($t=128$). \textbf{(d) OGBench Cube (Task \#1):} Online GD misses the target placement, whereas \method{} grasps and places the red cube onto the target ($t=14$).}
\label{fig:s_qualitative}
\end{figure*}

\begin{table*}[t]
\centering
\caption{\textbf{Success rate (\%) per model seed.} Tasks per seed: Push-T 250 (50 official + 200 held-out), others 50. Mean $\pm$ s.e.\ over seeds; pooled successes. MM, TT, TM: the three \method{} instantiations (\mmm{}/\ttt{}/\tmm{} at L3, \mmmm{}/\tttt{}/\tmmm{} at L4).}
\label{tab:s_seeds}
\footnotesize
\begin{tabular*}{\textwidth}{@{\extracolsep{\fill}}llccccc@{}}
\toprule
Setting & Planner & Seed 42 & Seed 43 & Seed 44 & Mean & Pooled \\
\midrule
Push-T L2 & Online GD & 40.0 & 44.8 & 37.6 & 40.8\se{2.1} & 306/750 \\
 & L1-only (GD$_{10}$ / MLP) & 38.8 & 37.2 & 37.2 & 37.7\se{0.5} & 283/750 \\
 & \mm & 56.4 & 52.4 & 58.8 & 55.9\se{1.9} & 419/750 \\
 & \ttarch & 65.6 & 64.0 & 60.0 & 63.2\se{1.7} & 474/750 \\
 & \tm & 66.4 & 66.8 & 60.8 & 64.7\se{1.9} & 485/750 \\
\addlinespace[3pt]
FourRoom L2 & Online GD & 90.0 & 84.0 & 76.0 & 83.3\se{4.1} & 125/150 \\
 & L1-only (GD$_{10}$ / MLP) & 78.0 & 78.0 & 72.0 & 76.0\se{2.0} & 114/150 \\
 & \mm & 92.0 & 90.0 & 94.0 & 92.0\se{1.2} & 138/150 \\
 & \ttarch & 100.0 & 100.0 & 100.0 & 100.0\se{0.0} & 150/150 \\
 & \tm & 100.0 & 100.0 & 100.0 & 100.0\se{0.0} & 150/150 \\
\addlinespace[3pt]
AntMaze L2 & Online GD & 46.0 & 50.0 & 40.0 & 45.3\se{2.9} & 68/150 \\
 & L1-only (GD$_{10}$ / MLP) & 72.0 & 80.0 & 66.0 & 72.7\se{4.1} & 109/150 \\
 & \mm & 82.0 & 72.0 & 82.0 & 78.7\se{3.3} & 118/150 \\
 & \ttarch & 86.0 & 86.0 & 62.0 & 78.0\se{8.0} & 117/150 \\
 & \tm & 84.0 & 80.0 & 72.0 & 78.7\se{3.5} & 118/150 \\
\addlinespace[3pt]
Cube L2 & Online GD & 40.0 & 54.0 & 50.0 & 48.0\se{4.2} & 72/150 \\
 & L1-only (GD$_{10}$ / MLP) & 36.0 & 70.0 & 44.0 & 50.0\se{10.3} & 75/150 \\
 & \mm & 86.0 & 76.0 & 80.0 & 80.7\se{2.9} & 121/150 \\
 & \ttarch & 74.0 & 80.0 & 72.0 & 75.3\se{2.4} & 113/150 \\
 & \tm & 74.0 & 78.0 & 70.0 & 74.0\se{2.3} & 111/150 \\
\addlinespace[3pt]
Push-T L3 & Online GD & 17.2 & 13.6 & 14.8 & 15.2\se{1.1} & 114/750 \\
 & \mmm & 22.0 & 14.0 & 13.2 & 16.4\se{2.8} & 123/750 \\
 & \ttt & 18.4 & 16.4 & 20.0 & 18.3\se{1.0} & 137/750 \\
 & \tmm & 20.4 & 15.2 & 18.0 & 17.9\se{1.5} & 134/750 \\
\addlinespace[3pt]
FourRoom L3 & Online GD & 98.0 & 88.0 & 90.0 & 92.0\se{3.1} & 138/150 \\
 & \mmm & 100.0 & 98.0 & 100.0 & 99.3\se{0.7} & 149/150 \\
 & \ttt & 100.0 & 100.0 & 100.0 & 100.0\se{0.0} & 150/150 \\
 & \tmm & 100.0 & 100.0 & 100.0 & 100.0\se{0.0} & 150/150 \\
\addlinespace[3pt]
AntMaze L3 & Online GD & 58.0 & 54.0 & 64.0 & 58.7\se{2.9} & 88/150 \\
 & \mmm & 92.0 & 74.0 & 76.0 & 80.7\se{5.7} & 121/150 \\
 & \ttt & 86.0 & 78.0 & 78.0 & 80.7\se{2.7} & 121/150 \\
 & \tmm & 90.0 & 74.0 & 78.0 & 80.7\se{4.8} & 121/150 \\
\addlinespace[3pt]
Push-T L4 & Online GD & 0.4 & 0.0 & 0.0 & 0.1\se{0.1} & 1/750 \\
 & \mmmm & 1.6 & 0.0 & 0.0 & 0.5\se{0.5} & 4/750 \\
 & \tttt & 0.0 & 0.0 & 0.0 & 0.0\se{0.0} & 0/750 \\
 & \tmmm & 3.6 & 0.0 & 0.0 & 1.2\se{1.2} & 9/750 \\
\addlinespace[3pt]
FourRoom L4 & Online GD & 86.0 & 96.0 & 92.0 & 91.3\se{2.9} & 137/150 \\
 & \mmmm & 90.0 & 96.0 & 96.0 & 94.0\se{2.0} & 141/150 \\
 & \tttt & 76.0 & 92.0 & 92.0 & 86.7\se{5.3} & 130/150 \\
 & \tmmm & 90.0 & 92.0 & 94.0 & 92.0\se{1.2} & 138/150 \\
\addlinespace[3pt]
AntMaze L4 & Online GD & 56.0 & 60.0 & 70.0 & 62.0\se{4.2} & 93/150 \\
 & \mmmm & 74.0 & 88.0 & 88.0 & 83.3\se{4.7} & 125/150 \\
 & \tttt & 90.0 & 88.0 & 88.0 & 88.7\se{0.7} & 133/150 \\
 & \tmmm & 90.0 & 76.0 & 86.0 & 84.0\se{4.2} & 126/150 \\
\bottomrule
\end{tabular*}
\end{table*}

\Cref{tab:s_seeds} gives the success rate of every planner for every model seed; Table~1  reports its mean and standard error. \Cref{tab:s_tests} lists the paired tests behind the comparisons in Sec.~5.2. On Push-T L4, H-JEPA's 4-level temporal abstraction assigns a 40-step top-level stride ($\Delta_4=40$, half of the 80-step episode), which aliases contact-rich T-block manipulations and causes all planners (including Online GD) to fail overall ($\le 1.2\%$, with seeds 43 and 44 yielding $0/250$ for every planner).

\Cref{tab:s_pusht_split} separates the official and held-out Push-T tasks of the L2 models. On both task sets, L1-only is at or below Online GD, \mm{} is 13--16\,pp above Online GD, and \tm{} and \ttarch{} are another 6--12\,pp above \mm{}. \Cref{tab:s_ladder} repeats the per-level amortization study of Table~2 (Online GD and MLP rows) on all four L2 benchmarks with three seeds. Amortizing only level~1 (with 10 warm-started top-level GD iterations) is $2.5$--$4.6\times$ faster than Online GD and changes success by $-7.3$ to $+27.3$\,pp; amortizing both levels is required for the full speed-up.

\begin{table}[t]
\centering
\caption{\textbf{Push-T L2: official \vs held-out tasks} (3 seeds). Held-out tasks: official procedure on the validation episodes, task seeds 1003--1006.}
\label{tab:s_pusht_split}
\footnotesize
\setlength{\tabcolsep}{3pt}
\begin{tabular*}{\columnwidth}{@{\extracolsep{\fill}}lcccc@{}}
\toprule
Planner & \shortstack{Official\\(\%)} & Succ. & \shortstack{Held-out\\(\%)} & Succ. \\
\midrule
Online GD & 40.7\se{5.2} & 61/150 & 40.8\se{1.4} & 245/600 \\
L1-only (GD$_{10}$ / MLP) & 35.3\se{1.8} & 53/150 & 38.3\se{0.4} & 230/600 \\
\mm & 54.0\se{2.0} & 81/150 & 56.3\se{1.9} & 338/600 \\
\ttarch & 66.0\se{3.1} & 99/150 & 62.5\se{1.4} & 375/600 \\
\tm & 64.0\se{3.1} & 96/150 & 64.8\se{1.7} & 389/600 \\
\bottomrule
\end{tabular*}
\end{table}

\begin{table*}[t]
\centering
\caption{\textbf{Paired differences for Table~1} (pp; exact two-sided McNemar $p$ over matched (seed, task) pairs in parentheses).}
\label{tab:s_tests}
\footnotesize
\setlength{\tabcolsep}{4pt}
\begin{tabular*}{\textwidth}{@{\extracolsep{\fill}}lccccc@{}}
\toprule
Setting & \shortstack{MM \\ vs.\ Online GD} & \shortstack{TT \\ vs.\ Online GD} & \shortstack{TM \\ vs.\ Online GD} & \shortstack{TM \\ vs.\ MM} & \shortstack{TT \\ vs.\ TM} \\
\midrule
Push-T L2 & +15.1 ($1.8{\times}10^{-11}$) & +22.4 ($1.4{\times}10^{-22}$) & +23.9 ($9.6{\times}10^{-24}$) & +8.8 ($1.7{\times}10^{-5}$) & -1.5 (0.46) \\
FourRoom L2 & +8.7 (0.041) & +16.7 ($6.0{\times}10^{-8}$) & +16.7 ($6.0{\times}10^{-8}$) & +8.0 ($4.9{\times}10^{-4}$) & +0.0 (1.00) \\
AntMaze L2 & +33.3 ($2.9{\times}10^{-9}$) & +32.7 ($3.2{\times}10^{-10}$) & +33.3 ($1.8{\times}10^{-10}$) & +0.0 (1.00) & -0.7 (1.00) \\
Cube L2 & +32.7 ($6.8{\times}10^{-10}$) & +27.3 ($1.3{\times}10^{-8}$) & +26.0 ($1.8{\times}10^{-6}$) & -6.7 (0.14) & +1.3 (0.83) \\
\addlinespace[3pt]
Push-T L3 & +1.2 (0.52) & +3.1 (0.086) & +2.7 (0.13) & +1.5 (0.42) & +0.4 (0.88) \\
FourRoom L3 & +7.3 (0.0034) & +8.0 ($4.9{\times}10^{-4}$) & +8.0 ($4.9{\times}10^{-4}$) & +0.7 (1.00) & +0.0 (1.00) \\
AntMaze L3 & +22.0 ($5.1{\times}10^{-5}$) & +22.0 ($6.7{\times}10^{-5}$) & +22.0 ($1.4{\times}10^{-4}$) & +0.0 (1.00) & +0.0 (1.00) \\
\addlinespace[3pt]
Push-T L4 & +0.4 (0.38) & -0.1 (1.00) & +1.1 (0.0078) & +0.7 (0.12) & -1.2 (0.0039) \\
FourRoom L4 & +2.7 (0.48) & -4.7 (0.26) & +0.7 (1.00) & -2.0 (0.55) & -5.3 (0.077) \\
AntMaze L4 & +21.3 ($7.7{\times}10^{-5}$) & +26.7 ($9.0{\times}10^{-8}$) & +22.0 ($1.9{\times}10^{-5}$) & +0.7 (1.00) & +4.7 (0.31) \\
\bottomrule
\end{tabular*}
\end{table*}

\begin{table*}[t]
\centering
\caption{\textbf{Per-level amortization on L2} (3 seeds; Push-T 750 tasks, others 150). L1-only: MLP at level 1, 10 warm-started GD iterations at the top level. Differences in pp (McNemar $p$); last column: L1-only speed-up over Online GD.}
\label{tab:s_ladder}
\footnotesize
\begin{tabular*}{\textwidth}{@{\extracolsep{\fill}}lcccccc@{}}
\toprule
Benchmark & L1-only vs.\ GD & \mm{} vs.\ GD & \mm{} vs.\ L1-only & \tm{} vs.\ \mm & \ttarch{} vs.\ \tm & L1-only speed-up \\
\midrule
Push-T & -3.1 (0.17) & +15.1 ($1.8{\times}10^{-11}$) & +18.1 ($1.8{\times}10^{-17}$) & +8.8 ($1.7{\times}10^{-5}$) & -1.5 (0.46) & $4.6\times$ \\
FourRoom & -7.3 (0.061) & +8.7 (0.041) & +16.0 ($1.2{\times}10^{-4}$) & +8.0 ($4.9{\times}10^{-4}$) & +0.0 (1.00) & $4.1\times$ \\
AntMaze & +27.3 ($9.6{\times}10^{-8}$) & +33.3 ($2.9{\times}10^{-9}$) & +6.0 (0.26) & +0.0 (1.00) & -0.7 (1.00) & $4.2\times$ \\
Cube & +2.0 (0.79) & +32.7 ($6.8{\times}10^{-10}$) & +30.7 ($6.9{\times}10^{-9}$) & -6.7 (0.14) & +1.3 (0.83) & $2.5\times$ \\
\bottomrule
\end{tabular*}
\end{table*}

\section{Ablation details}
\label{sec:s_ablation}

\Cref{tab:s_ablation} lists all variants of the Push-T ablation (seed 42, 250 paired tasks), including three that are not shown in the main text: the top-level MLP combined with level-1 GD whose bypass solves are capped at 10 iterations, and the level-1 \texttt{tf\_parallel} and \texttt{tf\_ar\_action} decoders under the MLP top level. Neither level-1 Transformer improves over the level-1 MLP ($+1.2$ and $0.0$\,pp).

\begin{table*}[t]
\centering
\caption{\textbf{All Push-T ablation variants} (L2, seed 42, 250 paired tasks). Differences in pp (McNemar $p$). Time: s/episode, first 10 official tasks, from three studies (main: Online GD, L1-only, \mm{}; mechanism: 4-layer; multi-benchmark: 1-layer); repeated runs differ by up to about 10\%.}
\label{tab:s_ablation}
\footnotesize
\begin{tabular*}{\textwidth}{@{\extracolsep{\fill}}llccccccc@{}}
\toprule
Top & L1 & Official & Held-out & Succ.\ (\%) & vs.\ \mm & vs.\ Online GD & Time (s) \\
\midrule
GD & GD & 20/50 & 80/200 & 40.0 & -16.4 ($5.5{\times}10^{-5}$) & -- & 25.27 \\
GD$_{10}$ & MLP & 19/50 & 78/200 & 38.8 & -17.6 ($4.9{\times}10^{-6}$) & -1.2 (0.84) & 5.55 \\
MLP & GD & 24/50 & 100/200 & 49.6 & -6.8 (0.040) & +9.6 (0.020) & 6.64 \\
MLP & GD (bypass 10) & 23/50 & 97/200 & 48.0 & -8.4 (0.014) & +8.0 (0.062) & 5.48 \\
MLP & MLP & 28/50 & 113/200 & 56.4 & -- & +16.4 ($5.5{\times}10^{-5}$) & 1.43 \\
\addlinespace[3pt]
\texttt{tf\_parallel} (4L) & MLP & 31/50 & 130/200 & 64.4 & +8.0 (0.022) & +24.4 ($2.7{\times}10^{-10}$) & 1.34 \\
\texttt{tf\_ar\_action} (4L) & MLP & 32/50 & 114/200 & 58.4 & +2.0 (0.63) & +18.4 ($4.7{\times}10^{-6}$) & 1.40 \\
\texttt{tf\_ar\_wm} (4L) & MLP & 32/50 & 137/200 & 67.6 & +11.2 (0.0020) & +27.6 ($4.2{\times}10^{-13}$) & 1.39 \\
\texttt{tf\_ar\_wm} (1L, \tm) & MLP & 34/50 & 132/200 & 66.4 & +10.0 (0.0035) & +26.4 ($3.6{\times}10^{-10}$) & 1.34 \\
\addlinespace[3pt]
MLP & \texttt{tf\_parallel} (4L) & 24/50 & 120/200 & 57.6 & +1.2 (0.78) & +17.6 ($2.8{\times}10^{-5}$) & 1.34 \\
MLP & \texttt{tf\_ar\_action} (4L) & 27/50 & 114/200 & 56.4 & +0.0 (1.00) & +16.4 ($1.4{\times}10^{-4}$) & 1.33 \\
MLP & \texttt{tf\_ar\_wm} (4L) & 24/50 & 110/200 & 53.6 & -2.8 (0.42) & +13.6 (0.0012) & 1.31 \\
\addlinespace[3pt]
\texttt{tf\_ar\_wm} (4L) & \texttt{tf\_ar\_wm} (4L) & 33/50 & 137/200 & 68.0 & +11.6 (0.0031) & +28.0 ($1.3{\times}10^{-10}$) & 1.43 \\
\texttt{tf\_ar\_wm} (1L) & \texttt{tf\_ar\_wm} (1L) & 34/50 & 130/200 & 65.6 & +9.2 (0.011) & +25.6 ($2.6{\times}10^{-10}$) & 1.37 \\
\bottomrule
\end{tabular*}
\end{table*}

\section{Timing}
\label{sec:s_timing}

\begin{table*}[t]
\centering
\caption{\textbf{Time per episode (s)} (seed 42, first 10 official tasks, sequential runs on one H200, including encoding and simulation). GD share: fraction of Online GD time spent in the planner (\%).}
\label{tab:s_time}
\footnotesize
\begin{tabular*}{\textwidth}{@{\extracolsep{\fill}}lccccccc@{}}
\toprule
Setting & Online GD & L1-only & MM & TT & TM & TM speed-up & GD share \\
\midrule
Push-T L2 & 25.27 & 5.55 & 1.43 & 1.37 & 1.34 & $18.8\times$ & 97.7 \\
FourRoom L2 & 45.08 & 10.89 & 2.76 & 2.93 & 2.93 & $15.4\times$ & 97.2 \\
AntMaze L2 & 78.87 & 18.75 & 5.23 & 5.70 & 5.69 & $13.9\times$ & 96.9 \\
Cube L2 & 3.44 & 1.37 & 1.18 & 1.16 & 1.19 & $2.9\times$ & 79.9 \\
\addlinespace[3pt]
Push-T L3 & 20.34 & -- & 1.49 & 1.49 & 1.48 & $13.7\times$ & 97.7 \\
FourRoom L3 & 36.61 & -- & 3.02 & 3.19 & 3.10 & $11.8\times$ & 96.6 \\
AntMaze L3 & 84.28 & -- & 6.09 & 6.28 & 6.23 & $13.5\times$ & 96.9 \\
\addlinespace[3pt]
Push-T L4 & 18.81 & -- & 1.91 & 1.96 & 1.91 & $9.9\times$ & 97.6 \\
FourRoom L4 & 46.22 & -- & 4.21 & 4.49 & 4.29 & $10.8\times$ & 96.9 \\
AntMaze L4 & 92.45 & -- & 7.12 & 7.30 & 7.15 & $12.9\times$ & 97.3 \\
\bottomrule
\end{tabular*}
\end{table*}

\begin{table*}[t]
\centering
\caption{\textbf{Solver time per level (s/episode).} Rest: encoding, simulation and code outside the per-level solvers.}
\label{tab:s_levels}
\footnotesize
\begin{tabular*}{\textwidth}{@{\extracolsep{\fill}}llrrrrrr@{}}
\toprule
Setting & Planner & Level 4 & Level 3 & Level 2 & Level 1 & Rest & Total \\
\midrule
Push-T L2 & Online GD & -- & -- & 17.22 & 7.31 & 0.74 & 25.27 \\
 & TM & -- & -- & 0.31 & 0.45 & 0.59 & 1.34 \\
\addlinespace[2pt]
FourRoom L2 & Online GD & -- & -- & 32.57 & 10.95 & 1.55 & 45.08 \\
 & TM & -- & -- & 0.83 & 0.64 & 1.46 & 2.93 \\
\addlinespace[2pt]
AntMaze L2 & Online GD & -- & -- & 47.64 & 28.33 & 2.91 & 78.87 \\
 & TM & -- & -- & 1.38 & 1.39 & 2.92 & 5.69 \\
\addlinespace[2pt]
Cube L2 & Online GD & -- & -- & 0.68 & 2.03 & 0.74 & 3.44 \\
 & TM & -- & -- & 0.02 & 0.40 & 0.76 & 1.19 \\
\addlinespace[2pt]
Push-T L3 & Online GD & -- & 7.49 & 5.17 & 7.09 & 0.59 & 20.34 \\
 & TM & -- & 0.16 & 0.31 & 0.44 & 0.57 & 1.48 \\
\addlinespace[2pt]
FourRoom L3 & Online GD & -- & 15.36 & 9.32 & 10.42 & 1.52 & 36.61 \\
 & TM & -- & 0.45 & 0.55 & 0.63 & 1.46 & 3.10 \\
\addlinespace[2pt]
AntMaze L3 & Online GD & -- & 32.78 & 19.91 & 28.44 & 3.15 & 84.28 \\
 & TM & -- & 0.81 & 1.06 & 1.40 & 2.96 & 6.23 \\
\addlinespace[2pt]
Push-T L4 & Online GD & 2.64 & 4.03 & 4.78 & 6.76 & 0.59 & 18.81 \\
 & TM & 0.08 & 0.50 & 0.32 & 0.44 & 0.57 & 1.91 \\
\addlinespace[2pt]
FourRoom L4 & Online GD & 10.28 & 9.97 & 10.91 & 13.30 & 1.77 & 46.22 \\
 & TM & 0.31 & 0.79 & 0.67 & 0.76 & 1.77 & 4.29 \\
\addlinespace[2pt]
AntMaze L4 & Online GD & 19.35 & 19.28 & 20.42 & 30.38 & 3.03 & 92.45 \\
 & TM & 0.50 & 1.53 & 1.06 & 1.09 & 2.97 & 7.15 \\
\bottomrule
\end{tabular*}
\end{table*}

\begin{table*}[t]
\centering
\caption{\textbf{Solves that still run gradient descent.} Percentage (count) of solves per level that are H-JEPA's end-of-episode bypass solves (no parent macro-action), where all \method{} instantiations run 10 GD iterations; all other solves use the proposers. Counts are identical for MM, TT and TM. Last column: mean Online GD iterations per solve, top level first.}
\label{tab:s_bypass}
\footnotesize
\begin{tabular*}{\textwidth}{@{\extracolsep{\fill}}lccccc@{}}
\toprule
Setting & Level 4 & Level 3 & Level 2 & Level 1 & Online GD iterations \\
\midrule
Push-T L2 & -- & -- & 0 (0/70) & 30 (30/100) & 39.3, 36.2 \\
FourRoom L2 & -- & -- & 0 (0/190) & 14 (30/220) & 36.8, 32.2 \\
AntMaze L2 & -- & -- & 0 (0/310) & 14 (50/360) & 33.8, 49.5 \\
Cube L2 & -- & -- & 0 (0/10) & 75 (30/40) & 32.1, 38.6 \\
\addlinespace[3pt]
Push-T L3 & -- & 0 (0/60) & 14 (10/70) & 30 (30/100) & 40.8, 35.7, 46.5 \\
FourRoom L3 & -- & 0 (0/170) & 11 (20/190) & 14 (30/220) & 33.7, 31.8, 31.2 \\
AntMaze L3 & -- & 0 (0/280) & 10 (30/310) & 14 (50/360) & 36.5, 33.2, 46.7 \\
\addlinespace[3pt]
Push-T L4 & 0 (0/40) & 33 (20/60) & 14 (10/70) & 30 (30/100) & 30.4, 30.6, 31.2, 42.1 \\
FourRoom L4 & 0 (0/140) & 18 (30/170) & 11 (20/190) & 14 (30/220) & 33.7, 32.4, 31.6, 33.8 \\
AntMaze L4 & 0 (0/230) & 21 (60/290) & 9 (30/320) & 11 (40/360) & 37.0, 34.8, 32.8, 51.8 \\
\bottomrule
\end{tabular*}
\end{table*}

\Cref{tab:s_time} reports the time per episode of every planner. The three \method{} instantiations (MM, TT, TM) run within 9\% of each other on every setting. Online GD spends 96.6--97.7\% of its episode time inside the planner on all settings except Cube (79.9\%). \Cref{tab:s_levels} splits the time by level, and \cref{tab:s_bypass} counts the solves that still run gradient descent under the \method{} instantiations: none at the top level, and 9--33\% of the subordinate solves (75\% of the level-1 solves on Cube).

\section{Proposers}
\label{sec:s_proposers}

\begin{table*}[t]
\centering
\caption{\textbf{Proposer parameter counts and training times across all benchmarks, depths, and architectures}: 30k AdamW steps, batch 256, learning rate $3\times10^{-4}$, 1k warm-up steps, cosine decay; top level $K=32$, $\lambda_{\mathrm{bc}}=0.05$; subordinate levels $K=16$, $\lambda_{\mathrm{bc}}=1$. Training: wall-clock time per run on one H200 (range over models).}
\label{tab:s_train}
\footnotesize
\begin{tabular*}{\textwidth}{@{\extracolsep{\fill}}llccccc@{}}
\toprule
Benchmark & Level & Architecture & Layers & Params (M) & Models & Training (h) \\
\midrule
Push-T & top & MLP & -- & 9.26--10.06 & 9 & 0.8--3.0 \\
 & top & \texttt{tf\_parallel} & 4 & 10.14 & 1 & 2.1 \\
 & top & \texttt{tf\_ar\_action} & 4 & 10.14 & 1 & 4.1 \\
 & top & \texttt{tf\_ar\_wm} & 1 & 3.43--3.44 & 9 & 1.0--3.6 \\
 & top & \texttt{tf\_ar\_wm} & 4 & 10.14 & 1 & 4.2 \\
 & subordinate & MLP & -- & 9.33--9.53 & 18 & 0.3--2.4 \\
 & subordinate & \texttt{tf\_parallel} & 4 & 9.61 & 1 & 0.8 \\
 & subordinate & \texttt{tf\_ar\_action} & 4 & 9.61 & 1 & 1.2 \\
 & subordinate & \texttt{tf\_ar\_wm} & 1 & 3.50--3.51 & 18 & 0.3--2.9 \\
 & subordinate & \texttt{tf\_ar\_wm} & 4 & 9.61 & 1 & 1.2 \\
\addlinespace[3pt]
FourRoom & top & MLP & -- & 9.39--10.58 & 9 & 0.5--1.4 \\
 & top & \texttt{tf\_ar\_wm} & 1 & 3.43--3.44 & 9 & 0.8--2.1 \\
 & subordinate & MLP & -- & 9.33--9.53 & 18 & 0.2--1.1 \\
 & subordinate & \texttt{tf\_ar\_wm} & 1 & 3.50--3.51 & 18 & 0.3--1.6 \\
\addlinespace[3pt]
AntMaze & top & MLP & -- & 9.99--10.78 & 9 & 0.7--1.3 \\
 & top & \texttt{tf\_ar\_wm} & 1 & 3.73 & 9 & 1.0--1.9 \\
 & subordinate & MLP & -- & 10.12--11.30 & 18 & 0.2--0.7 \\
 & subordinate & \texttt{tf\_ar\_wm} & 1 & 3.87--3.94 & 18 & 0.3--0.7 \\
\addlinespace[3pt]
Cube & top & MLP & -- & 9.26 & 3 & 0.7--0.8 \\
 & top & \texttt{tf\_ar\_wm} & 1 & 3.43 & 3 & 0.6 \\
 & subordinate & MLP & -- & 10.02 & 3 & 0.7 \\
 & subordinate & \texttt{tf\_ar\_wm} & 1 & 3.54 & 3 & 0.3 \\
\bottomrule
\end{tabular*}
\end{table*}

\paragraph{Architectures.}
The MLP proposer has four pre-LayerNorm residual blocks of width 1024 and predicts all $K_l \times H_l$ actions in one pass. The Transformer proposers have one pre-LayerNorm decoder layer ($d_{\text{model}}=384$, 6 heads, feed-forward width 1536) with causal self-attention over the step queries and cross-attention to context tokens (the start latents, the goal latent and its offset from the start, the horizon at the top level or the parent macro-action at subordinate levels, and the candidate embedding). \texttt{tf\_parallel} builds every step query from the start latent and a zero action, \texttt{tf\_ar\_action} additionally feeds back the previous macro-action, and \texttt{tf\_ar\_wm} also feeds back the latent state predicted by the frozen predictor for the previous macro-action (Sec.~4.3). The architecture ablation study (Table~2) uses 4-layer variants of all three Transformers with 10.14M (top level) and 9.61M (level 1) parameters.

\paragraph{Training.}
Proposers are trained on latents of the H-JEPA training split computed with the frozen encoders. Each proposer is trained for 30k AdamW steps with batch size 256, learning rate $3\times10^{-4}$, 1k linear warm-up steps and cosine decay. The relaxed winner-take-all cost uses $\varepsilon=0.05$; the behavior-cloning weight is $\lambda_{\mathrm{bc}}=0.05$ for the top-level proposers ($K_L=32$) and $\lambda_{\mathrm{bc}}=1$ for the subordinate proposers ($K_l=16$). \Cref{tab:s_train} lists sizes and training times.

\paragraph{Inference.}
The candidates of a proposer are scored together with a zero-action anchor and, at the top level, the shifted previous plan, in one batched rollout of the frozen predictor (for \texttt{tf\_ar\_wm}, the rollout computed during generation). H-JEPA's end-of-episode bypass solves, which skip the upper levels and have no parent macro-action, keep gradient descent with 10 iterations (\cref{tab:s_bypass}). \Cref{tab:s_micro} breaks down the latency of one top-level call.

\begin{table*}[t]
\centering
\caption{\textbf{Latency of one top-level call} (ms, mean; Push-T, seed 42, $H_2=8$, 32 candidates, 4-layer Transformers). Proposer: forward pass; in-loop rollout: predictor calls inside the autoregressive loop; scoring: rollout and cost for selection (\texttt{tf\_ar\_wm} reuses its in-loop rollout).}
\label{tab:s_micro}
\footnotesize
\begin{tabular*}{\textwidth}{@{\extracolsep{\fill}}lccccc@{}}
\toprule
Top proposer & \shortstack{Params\\(M)} & Prop. & \shortstack{In-loop\\rollout} & Scoring & Total \\
\midrule
MLP & 10.06 & 0.57 & 0.00 & 29.96 & 30.53 \\
\texttt{tf\_parallel} (4L) & 10.14 & 4.95 & 0.00 & 30.06 & 35.24 \\
\texttt{tf\_ar\_action} (4L) & 10.14 & 25.89 & 0.00 & 29.79 & 56.30 \\
\texttt{tf\_ar\_wm} (4L) & 10.14 & 26.38 & 30.73 & 0.14 & 57.89 \\
\texttt{tf\_ar\_wm} (1L, \tm) & 3.44 & 10.39 & 30.95 & 0.13 & 41.47 \\
\bottomrule
\end{tabular*}
\end{table*}

\Cref{tab:s_mechanism} reports the complete 18-variant top-level mechanism study behind Table~3 and Fig.~3b (evaluated on the released H-JEPA Push-T L2 checkpoint and timed on an RTX 4090, whereas Tables~1, 2, and S2--S9 evaluate the self-trained H200 checkpoints), including offline terminal cost $\mathcal{C}_2$ (median and mean), first-step and full-horizon macro-action MSE on 192 validation segments ($H_2=8$), and closed-loop evaluations across the 150-task exploratory suite, the 200-task held-out confirmatory suite, and the 350-task pooled total. Notably, appending $R=2$ or $R=5$ online gradient refinement steps to the $K=32$ learned proposals monotonically decreases offline world-model cost $\mathcal{C}_2$ ($73.7 \to 50.2 \to 23.2$) while leaving first-macro action MSE nearly unchanged ($1.25 \to 1.12 \to 1.17$). However, in closed-loop execution, the small $+4.0$\,pp difference for $R=5$ on the 150-task exploratory split ($52.0\%$ \vs $48.0\%$, $p=0.44$; whereas $R=2$ drops to $42.0\%$, $p=0.30$) reverses on the 200-task confirmatory split ($49.5\%$ \vs $53.5\%$, $-4.0$\,pp, $p=0.38$), yielding virtually identical pooled accuracy across all 350 tasks ($50.6\%$ \vs $51.1\%$, $-0.6$\,pp, $p=0.91$) at $2.6\times$ higher episode latency ($5.96$\,s \vs $2.31$\,s). Once best-of-$K$ selection among $\lambda_{\mathrm{bc}}$-regularized proposals identifies a feasible topological mode basin, further unregularized local gradient steps primarily exploit local curvature of the frozen predictor $P_2$ rather than improving physical execution, while receding-horizon replanning already corrects small local tracking errors.

\begin{table*}[t]
\centering
\caption{\textbf{Complete top-level mechanism study on Push-T L2} (released H-JEPA checkpoint, seed 42; level 1 is the learned MLP except for Online GD). \emph{Offline}: terminal predictor cost $\mathcal{C}_2$ (median and mean) and macro-action MSE (first step and all 8 steps, prior-std units) over 192 validation segments ($H_2=8$). \emph{Closed loop} (success, \%): exploratory 150-task suite (official 50 + held-out seeds 1001--1002) with the difference to \mm{} in pp (exact McNemar $p$), confirmatory 200-task suite (held-out seeds 1003--1006), pooled 350 tasks; time per episode (s, RTX 4090).}
\label{tab:s_mechanism}
\footnotesize
\begin{tabular*}{\textwidth}{@{\extracolsep{\fill}}lrrrrrrrrr@{}}
\toprule
 & \multicolumn{4}{c}{Offline} & \multicolumn{5}{c}{Closed loop} \\ \cmidrule(lr){2-5}\cmidrule(l){6-10}Top-level solver / variant & \shortstack{Med.\\$\mathcal{C}_2$} & \shortstack{Mean\\$\mathcal{C}_2$} & \shortstack{MSE\\1st} & \shortstack{MSE\\all} & \shortstack{Expl.\\150} & \shortstack{vs.\ \mm{}\\(pp, $p$)} & \shortstack{Conf.\\200} & \shortstack{Pooled\\350} & \shortstack{Time\\(s)} \\
\midrule
\multicolumn{10}{@{}l}{\textit{Reference and search baselines}} \\
Expert (ground truth) & 215.0 & 216.1 & 0.00 & 0.00 & -- & -- & -- & -- & -- \\
Prior mean (zero action) & 372.7 & 391.4 & 3.96 & 3.79 & -- & -- & -- & -- & -- \\
Random shoot, $K=32$ & 275.0 & 270.7 & 6.74 & 7.22 & 0.0 & $-48.0$ ($4.2{\times}10^{-22}$) & -- & -- & 2.33 \\
Random shoot, $K=400$ & 218.0 & 206.5 & 6.53 & 7.76 & -- & -- & -- & -- & -- \\
Online GD, 10 it.\ (L1: MLP) & 52.3 & 54.7 & 4.52 & 7.14 & 34.7 & $-13.3$ (0.017) & -- & -- & 12.49 \\
Online GD, 30 it. & 9.3 & 10.7 & 4.23 & 6.45 & -- & -- & -- & -- & -- \\
Online GD, 90 it. & 2.5 & 2.9 & 4.22 & 6.92 & -- & -- & -- & -- & -- \\
Online GD, 30--90 it.\ (early stop) & 16.9 & 17.5 & 4.23 & 6.45 & 42.0 & $-6.0$ (0.30) & 37.5 & 39.4 & 44.32 \\
\addlinespace[3pt]
\multicolumn{10}{@{}l}{\textit{MLP top-level proposer, $\lambda_{\mathrm{bc}}=0.05$}} \\
MeanHead (average of 32, no WM) & 314.8 & 285.8 & 0.97 & 2.53 & 28.0 & $-20.0$ ($2.9{\times}10^{-4}$) & -- & -- & 2.21 \\
Best-of-$K$, $K=1$ & 278.9 & 255.9 & 1.20 & 3.21 & 18.7 & $-29.3$ ($3.6{\times}10^{-8}$) & -- & -- & 2.25 \\
Best-of-$K$, $K=2$ & 224.0 & 221.7 & 1.12 & 3.28 & -- & -- & -- & -- & -- \\
Best-of-$K$, $K=4$ & 165.3 & 182.3 & 0.91 & 3.49 & 38.7 & $-9.3$ (0.070) & -- & -- & 2.28 \\
Best-of-$K$, $K=8$ & 130.2 & 155.8 & 0.89 & 3.63 & 42.0 & $-6.0$ (0.27) & -- & -- & 2.34 \\
Best-of-$K$, $K=16$ & 87.3 & 135.5 & 1.07 & 3.84 & -- & -- & -- & -- & -- \\
\rowcolor{oursrow}Best-of-$K$, $K=32$ (\mm, $R=0$) & 73.7 & 112.6 & 1.25 & 3.98 & 48.0 & ref. & 53.5 & 51.1 & 2.31 \\
\addlinespace[3pt]
\multicolumn{10}{@{}l}{\textit{Variants of \mm{} ($K=32$)}} \\
$\lambda_{\mathrm{bc}}=0$ & 74.7 & 113.1 & 3.14 & 6.45 & 42.7 & $-5.3$ (0.30) & -- & -- & 2.26 \\
$+2$ GD steps ($R=2$) & 50.2 & 79.5 & 1.12 & 3.74 & 42.0 & $-6.0$ (0.30) & -- & -- & 3.76 \\
$+5$ GD steps ($R=5$) & 23.2 & 46.8 & 1.17 & 3.71 & 52.0 & $+4.0$ (0.44) & 49.5 & 50.6 & 5.96 \\
\bottomrule
\end{tabular*}
\end{table*}

\end{document}